\documentclass{article} 
\usepackage{booktabs} 
\usepackage{iclr2025_conference,times}

\usepackage{amsmath,amsfonts,bm}

\def\eqref#1{equation~\ref{#1}}

\def\1{\bm{1}}

\DeclareMathAlphabet{\mathsfit}{\encodingdefault}{\sfdefault}{m}{sl}
\SetMathAlphabet{\mathsfit}{bold}{\encodingdefault}{\sfdefault}{bx}{n}

\usepackage{listings}
\usepackage{xcolor}

\definecolor{codegreen}{rgb}{0,0.6,0}       
\definecolor{codegray}{rgb}{0.5,0.5,0.5}   
\definecolor{codepurple}{rgb}{0.58,0,0.82} 
\definecolor{backcolour}{rgb}{0,0,0}       
\definecolor{textcolor}{rgb}{1,1,1}        

\lstdefinestyle{mystyle}{
    backgroundcolor=\color{backcolour},   
    commentstyle=\color{codegreen},      
    keywordstyle=\color{magenta},        
    numberstyle=\tiny\color{codegray},   
    stringstyle=\color{codepurple},      
    basicstyle=\ttfamily\footnotesize\color{textcolor}, 
    breakatwhitespace=false,             
    breaklines=true,                     
    captionpos=b,                        
    keepspaces=true,                     
    numbers=left,                        
    numbersep=5pt,                       
    showspaces=false,                    
    showstringspaces=false,              
    showtabs=false,                      
    tabsize=2                            
}

\usepackage{tcolorbox}
\usepackage{url}
\usepackage{graphicx} 
\usepackage{booktabs}
\usepackage{times}
\usepackage{epsfig}
\usepackage{amsmath}
\usepackage{amsfonts}
\usepackage{cancel}
\usepackage[misc]{ifsym}

\usepackage{bbm}
\usepackage{mathtools}
\usepackage{amssymb}
\usepackage{pifont}
\usepackage{booktabs}
\usepackage{arydshln}
\usepackage{color}
\usepackage{colortbl}
\usepackage{subcaption}
\usepackage{multirow}
\usepackage{float}
\usepackage{rotfloat}
\usepackage{capt-of}
\usepackage{longtable}
\usepackage{diagbox}
\usepackage{makecell}
\usepackage{pgfplots}
\usepackage{xspace}
\usepackage{wrapfig}
\usepackage{enumitem}
\usepackage{fontawesome}
\usepackage{epigraph}
\usepackage{multirow}
\usepackage{multicol}
\usepackage{rotating}
\usepackage{amssymb}
\usepackage{pifont}
\usepackage{algorithm}
\usepackage{algpseudocode}
\usepackage{colortbl}
\definecolor{mycolor_blue}{HTML}{E7EFFA}
\definecolor{mycolor_green}{HTML}{E6F8E0}
\definecolor{mycolor_gray}{HTML}{ECECEC}
\definecolor{pearDark}{HTML}{2980B9}

\usepackage{tcolorbox}
\usepackage{fbox}
\definecolor{textcolor1}{rgb}{0.25,0.5,0.5}
\definecolor{textcolor2}{rgb}{0.7,0.25,0.25}
\definecolor{linkc}{rgb}{0, 0.44, 0.74}
\definecolor{eqc}{rgb}{1, 0, 0}
\definecolor{myy}{RGB}{126,95,0}
\definecolor{mygray}{gray}{.9}
\definecolor{bblue}{RGB}{30,80,120}
\definecolor{mygray1}{gray}{.7}
\definecolor{ggray}{RGB}{127,127,127}
\definecolor{mygreen}{RGB}{93,174,86}
\definecolor{scolor}{RGB}{111,168,220}
\definecolor{hcolor}{RGB}{111,176,81}
\definecolor{ocolor}{RGB}{224,103,102}
\definecolor{wcolor}{RGB}{246,178,107}
\definecolor{citecolor}{HTML}{229954}

\usepackage[pagebackref=false,breaklinks=true,colorlinks=True,urlcolor=eqc,citecolor=citecolor,linkcolor=eqc,bookmarks=false,urlcolor=scolor]{hyperref}

\title{PP-DocBee: Improving Multimodal Document Understanding Through a Bag of Tricks}

\author{Feng Ni$^\dag$\quad Kui Huang$^\dag$\quad Yao Lu$^\dag$\quad Wenyu Lv$^\dag$\quad Guanzhong Wang\quad Zeyu Chen\quad Yi Liu\\
PaddlePaddle Team, Baidu Inc. \\ 
\small \href{mailto:nifeng03@baidu.com}{nifeng03@@baidu.com} \quad 
\small \href{mailto:huangkui01@baidu.com}{huangkui01@baidu.com} \quad 
\small \href{mailto:lvwenyu01@baidu.com}{lvwenyu01@baidu.com} \quad
\small \href{mailto:liuyi22@baidu.com}{liuyi22@baidu.com}
}

\iclrfinalcopy 
\begin{document}

\def\thefootnote{$\dagger$}\footnotetext{Equal Contribution. }

\maketitle

\begin{abstract}

With the rapid advancement of digitalization, various document images are being applied more extensively in production and daily life, and there is an increasingly urgent need for fast and accurate parsing of the content in document images. Therefore, this report presents PP-DocBee, a novel multimodal large language model designed for end-to-end document image understanding. First, we develop a data synthesis strategy tailored to document scenarios in which we build a diverse dataset to improve the model generalization. Then, we apply a few training techniques, including dynamic proportional sampling, data preprocessing, and OCR postprocessing strategies. Extensive evaluations demonstrate the superior performance of PP-DocBee, achieving state-of-the-art results on English document understanding benchmarks and even outperforming existing open source and commercial models in Chinese document understanding. The source code and pre-trained models are available at \href{https://github.com/PaddlePaddle/PaddleMIX}{https://github.com/PaddlePaddle/PaddleMIX}.

\end{abstract}

\section{Introduction}
\label{sec:intro}

Recent advances in multimodal large language models (MLLMs)~\citep{llava,llava1.5,minigpt4,qwenvl} have demonstrated remarkable capabilities in general vision-language understanding through alignment of visual encoders~\citep{vit2021, clip} with Large Language Models (LLMs)~\citep{llama, vicuna,qwen}. Specifically, ViT is tasked with image processing to extract visual features. These features are subsequently processed and integrated by the MLP. The LLM component is responsible for understanding and generating text. This synergistic combination enables the model to process image and text information concurrently, facilitating the understanding of comprehensive multimodal documents.

Popular open source MLLM models such as Qwen2-VL~\citep{wang2024qwen2vl} and InternVL2~\citep{chen2024internvl2} follow the ``ViT+MLP+LLM'' paradigm and score high on the leaderboards of some document understanding tasks. However, these models exhibit significant limitations when processing text-rich visual content such as documents, tables, and charts~\citep{llmocr}, mainly due to two inherent constraints: (1) existing vision-to-text modules are optimized for natural image features rather than textual/structural representations, and (2) the current document understanding model has weak ability in Chinese scenarios.

Document image understanding has excellent potential in various fields. For example, corporate workflows require automatic extraction of financial data and contract terms, academic research requires efficient analysis of various articles and archives, and individuals must also process some receipts/forms intelligently. Although traditional approaches~\citep{li2022paddleocr} that combine optical character recognition (OCR) with rule-based systems have partially succeeded, they struggle with complex document layouts, poor image quality, and contextual reasoning requirements.

To effectively address these challenges, we introduce PP-DocBee, with the two main contributions elucidated below:
\begin{itemize}
    \item We propose a document data synthesis strategy that generates a 477k high-quality dataset on Chinese document understanding named PPInfinityDocData. We also collected a public dataset of 3.3M from various document sources.
    \item We validate the effectiveness of the data strategy on an open-source multimodal large model. The experiments demonstrate that the generic model can improve document understanding through a few techniques with our data. It achieves state-of-the-art performance, surpasses its counterparts, and maintains a lightweight and fast architecture.
\end{itemize}

\section{Data Synthesis Strategy}

\label{sec: method}


In terms of data quality, existing open-source document datasets have significant deficiencies. The lack of Chinese corpora, uneven quality of images and texts, absence of information extraction capabilities, and insufficient scene diversity are particularly prominent. Existing multimodal large models face multiple limitations while generating question-answer pairs regarding data generation. These
limitations include systemic defects such as uncontrollable generation costs, poor answer accuracy, and deviation from the focus of the question. These problems directly contribute to the inability to understand multimodal large language models (MLLMs) for Chinese documents.

\begin{figure}[h]
    \centering
    \includegraphics[width=12cm]{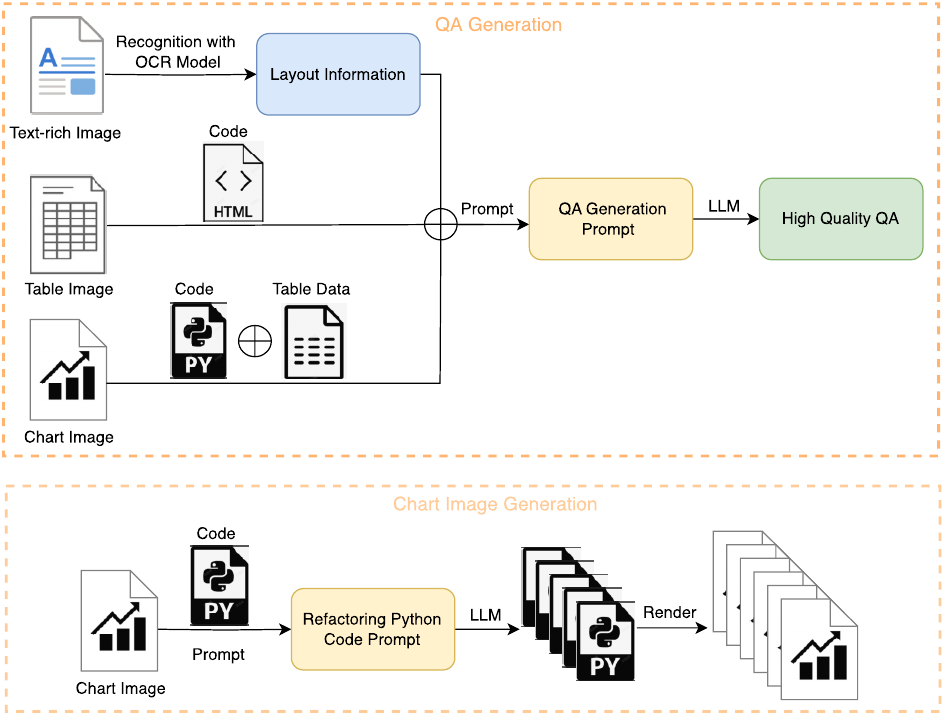}
    \caption{Document image QA generation and Chart image generation pipeline.}
    \label{fig:doc}
\end{figure}

We propose a data generation pipeline for document understanding to improve the Chinese semantic understanding and expand the coverage of multimodal scenarios, as shown in Figure~\ref{fig:doc}. This pipeline targets three typical document types: text-rich documents, tables, and charts, and designs differentiated data generation pipelines. A multimodal collaborative generation mechanism balances data quality and generation efficiency. The core technical strategies include: (1) constructing a cascaded processing architecture that combines a small Optical Character Recognition (OCR) model and a large language model, which can collaborate to make up for shortcomings; (2) developing a rendering engine based on semantic control, which helps expand Chinese chart images; and (3) designing document type-sensitive prompt engineering templates, which helps generate high-quality Chinese document question-answer(QA) pairs.

\subsection{Text-rich Document}
Text-rich documents are a common document type that contains a large amount of text and a small amount of pictures, tables, and charts. This data type helps the model understand the overall semantics of the document and general questions. 

\begin{figure}[h]
    \centering
    \includegraphics[width=\textwidth]{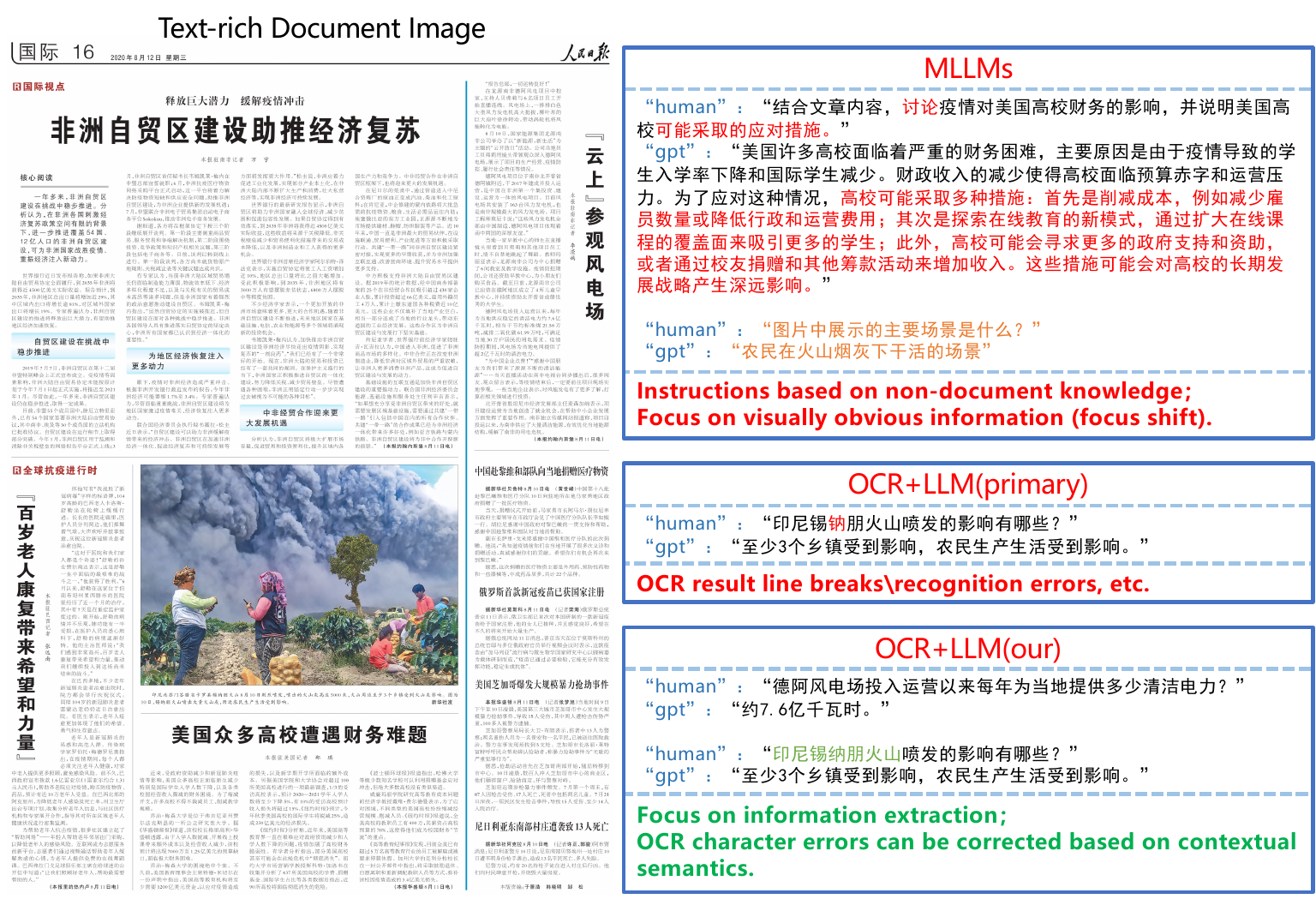}
    \caption{Text-rich document QA Generation. Red and orange indicate errors and inappropriate generation, and green indicates generation with fixed issues. The image comes from the open source dataset M6Doc~\cite{Cheng_2023_CVPR}. }
    \label{fig:text-rich}
\end{figure}

We compared the strategy of generating QA based on MLLMs directly from images with the strategy of generating QA based on an LLM in combination with a smaller OCR model. The strategy that uses MLLMs for generating QA can lead to knowledge expansion when generating instructions, which involves issues like generating instructions based on non-document knowledge. Additionally, this strategy often focuses on visually perceptible information and unconventional OCR-related questions. We provide an example of the text-rich document image and QA pairs generation shown in Figure~\ref{fig:text-rich}.

\textbf{Text-rich Images}. We filter complex-layout Chinese PDF documents from professional fields such as academic papers, financial reports, and research reports. Using document parsing tools, we construct a single-page document image dataset that includes mixed elements such as text, images, and formulas, preserving the layout features of the original documents. \\
\textbf{QA Generation}. We propose an OCR-LLM collaborative verification mechanism, which utilizes the PaddleOCR~\cite{li2022paddleocr} to extract accurate layout structures and textual information. Subsequently, the OCR outputs are integrated with the semantic understanding capabilities of a Large Language Model (ERNIE-Bot 4.0): 1) parse document images and obtain the corresponding layout information \texttt{'}\{json\_string\}\texttt{'} by using PaddleOCR, and the images are converted into text information. 2) design appropriate prompts to enable the LLM to correct OCR recognition errors based on the context semantics; 3) control the distribution of generated question-answer pairs through instruction templates. The prompt used is as follows: \\

\begin{tcolorbox}[colback=white,colframe=black,title=Prompt for Text-rich Document QA Generation]
question = f\texttt{"}You are a document data visual question-answering dialogue generation system. Your main task is to design instructions and corresponding answers based on the OCR layout information of the provided document image, so that when the obtained multimodal data is used for model training, the model can fully learn the multimodal document understanding ability.

Here is the OCR layout information extracted from a research report document image: \{json\_string\}. Please imagine that you are looking at the corresponding image instead of a simple json string based on the content (text-related characters are recognized by line, please fully splice and understand it), and then combine the information in the image to design instructions and answers from the perspective of professional research report readers.

When generating these dialogues, please make sure to follow the following guidelines:

These instructions must meet the following requirements:

1. Instructions focus on the ability to extract information from document images, and the answers can be directly observed from the image.

2. Instructions and answers must be highly relevant to the image, and instructions cannot be answered without the image. If the instruction provides too much information from the image, so that the instruction can be answered without the image, it is strictly prohibited.

3. The instructions must be generated based on the information in the image, and the accurate answer can be obtained in combination with the image content.

4. The text characters in the provided OCR layout information contain the exact answer to the answer. Please strictly ensure that the answer is correct, otherwise do not generate the instruction and answer.

5. For each layout area type (printed text, tables, charts, printed formulas, seals), if the document contains this type of information, then please generate instructions based on this type of content, otherwise no need to generate.

6. The answer to the instruction should be as concise and accurate as possible. Do not repeat the question and reply directly to the answer. At the same time, ensure that the answer is directly obtained from the original text of the image, and do not summarize.

7. Instructions and questions should not contain information about the layout structure.

8. Instructions should directly give questions, and do not use words such as \texttt{'}Please ask\texttt{'}, \texttt{'}Please answer\texttt{'}, and \texttt{'}In the document\texttt{'}.

9. When generating instructions related to tables, if the table contains relevant information such as units, percentages, positive and negative signs, please ensure the completeness of the answer.

Please generate at least 5 Chinese instructions and answers. You need to provide the generated content in JSON format. Please make sure that \texttt{```json```} is included in the output. You can refer to the following sample to organize your output: \{template\}.
\texttt{"}
\end{tcolorbox}



\subsection{Chart}

\begin{figure}[h]
    \centering
    \includegraphics[width=\textwidth]{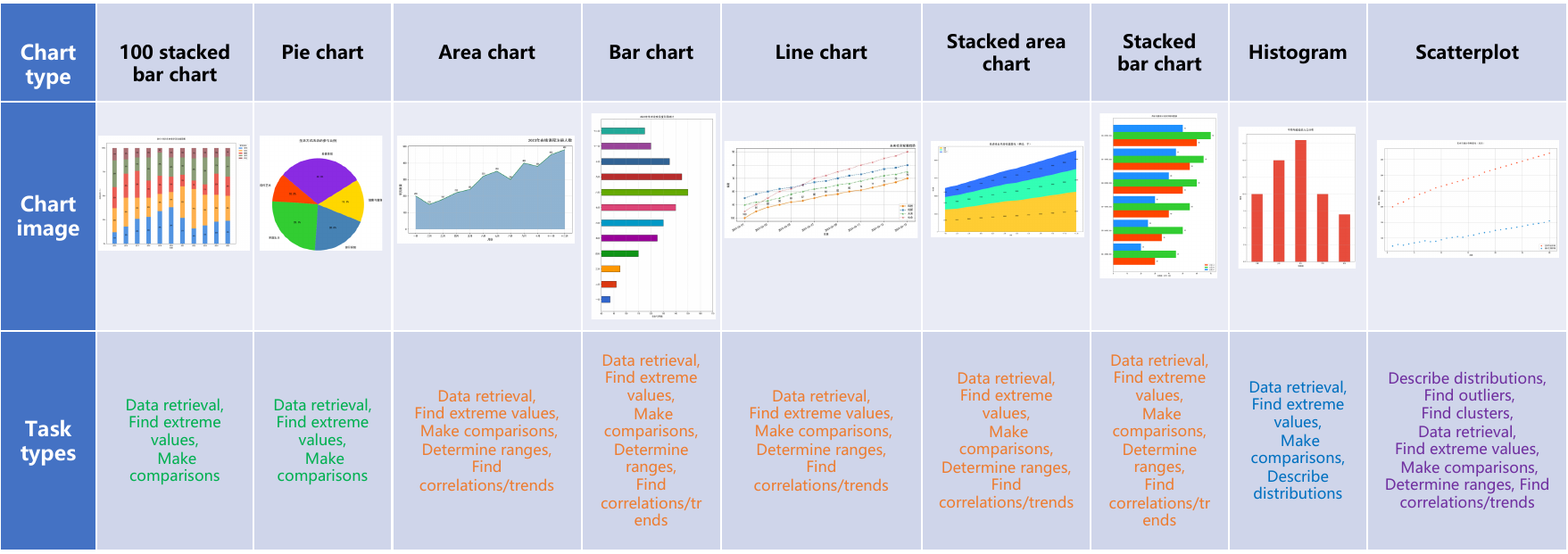}
    \caption{Common chart types and the corresponding task types }
    \label{fig:chart_type}
\end{figure}

For chart data, MMLMs generally perform poorly in analyzing charts, and there are fewer high-quality Chinese chart images. So we summarize the currently common nine types of chart types as Figure~\ref{fig:chart_type}, and design an image generation and QA generation scheme for each type of chart, and summarize the corresponding task types that can be asked.

\textbf{Chart Images}. We obtain 2,159 high-quality chart data(including images, codes and data tables) as our seed data by manual screening and filtering out incomplete image information and data with poor text-image relevance from the English chart dataset open-sourced by LLaVA-Chart~\cite{li2022paddleocr}, and develop a parametric Chinese chart image generation engine: 1) use LLM to semantically modify parameters in code corresponding to the chart (e.g., values, axes, colors, legends, themes), and randomly select and change the theme of the chart from the topics pool (e.g., "Art \& Design", "Science \& Nature"); 2) translate the text presented on the image into Chinese, when the LLM outputs code, while keeping other parameters unchanged. 3) prevent rendering issues such as image annotations exceeding the image range when the generated code is rendered into an image by fixing the code format; 4) parse the generated text into Python code and then render it into an image using Matplotlib. In this way, we obtain many Chinese chart images with rich scenarios. The prompt used is as follows, area chart for example:\\

\begin{tcolorbox}[colback=white,colframe=black,title=Prompt for Chart Image Generation (area chart)]
question = f\texttt{"}You are a highly intelligent AI familiar with data visualization and matplotlib.

Given the following matplotlib code of an area chart: \{code\}. Please generate a diversified version of this matplotlib code of area chart. Here are some options you should follow:\

1. Generate data points that fit the chart and the number of data points should be as much as you can to provide, but you should not skip any data point in your code. Do not add any comment in your code.\

2. Add or change some data point with new corresponding values to enrich the visualization and print the final table data you use in the code within triple backticks \texttt{(```)}, and don't include things like csv, plaintext in the triple backticks \texttt{(```)}.\

3. Modify the color scheme using specific color codes (e.g., \#RRGGBB) for better clarity or visual appeal. Avoid using color categories.\

4. Change width and height of the chart reasonably.\

5. Change the topic, headline, and data type (which fit the topic) of the chart, put the headline in an appropriate place that does not overlap with other things (make sure the headline does not overlap with the legend! Put these two things away), you can refer and choose one (not all) of the topics from: \{topics\_pool\}, but reduce using global topic, do not use temperature.\

6. Generate only an area chart with one column of value, not a stacked area chart.\

7. Assign annotation/text label on the chart. Do not use random data.\

You should choose some of these options, not all of them, to diversify the visualization.\

Different data points should have different values.\

You should give FULL code with ALL data points and don't miss any detail.\

Make sure the legend appears completely in the chart after the code is rendered.\

Print table data first in Chinese, the table data format should be able to directly write in csv file, then print your code (keep the topic, headline, and data type (which fit the topic) of the chart in Chinese).\
Include the code with  \texttt{``` ```} format.\texttt{"}
\end{tcolorbox}

\textbf{QA Generation}. With the help of the code and data tables corresponding to the chart image, QA generation will become more accurate. We design a data-chart driven question-answer generation framework: 1) extract statistical features from the code and data tables as question material; 2) match preset question templates based on chart types and task types as Figure~\ref{fig:chart_type}; 3) use LLM for semantic expansion and logical verification to generate QA pairs in Chinese that include professional questions and exact answers with the format as the template. The prompt used is as follows: \\

\begin{tcolorbox}[colback=white,colframe=black,title=Prompt for Chart QA Generation]
question = f\texttt{"}You are a highly intelligent AI familiar with data visualization and \{chart\_type\}.

Below is the matplotlib code for the \{chart\_type\} chart: \{code\} and the corresponding table data: \{table\_data\}.

Please imagine that you are looking at the image generated by the code, not the code itself.

Please generate questions of different task types based on the content of the chart.

The task type is \{task\_types\}.

Remember that in your answer, only the image of the chart is given, and your answer is based on the image. The table data is the real value of the relevant numerical value in the image, so make sure the answer is correct.

The value and label of the question are the real basis of your question, so make sure the answer is correct.

Avoid using invalid escape characters in strings.

Use approximate color names instead of hexadecimal colors.
If there are units, \% and other information in the chart, please ensure the integrity of the units , \% and other information in the answer.

In addition, I hope to save your output as a json file, so I hope you can organize your answer like \{template\}, and make sure the values of human and gpt in json are all in Chinese.
\texttt{"}
\end{tcolorbox}

\subsection{Table}

In addition to chart data, table data is important in document understanding. Similar to charts, high-quality Chinese table data suitable for multimodal large model training is also lacking. Therefore, we have designed a high-quality QA production strategy based on the internal table data used for layout analysis tasks (including table data and HTML code). We do not produce many separate data because the text-rich data will contain some table data.

\begin{figure}[h]
    \centering
    \includegraphics[width=11cm]{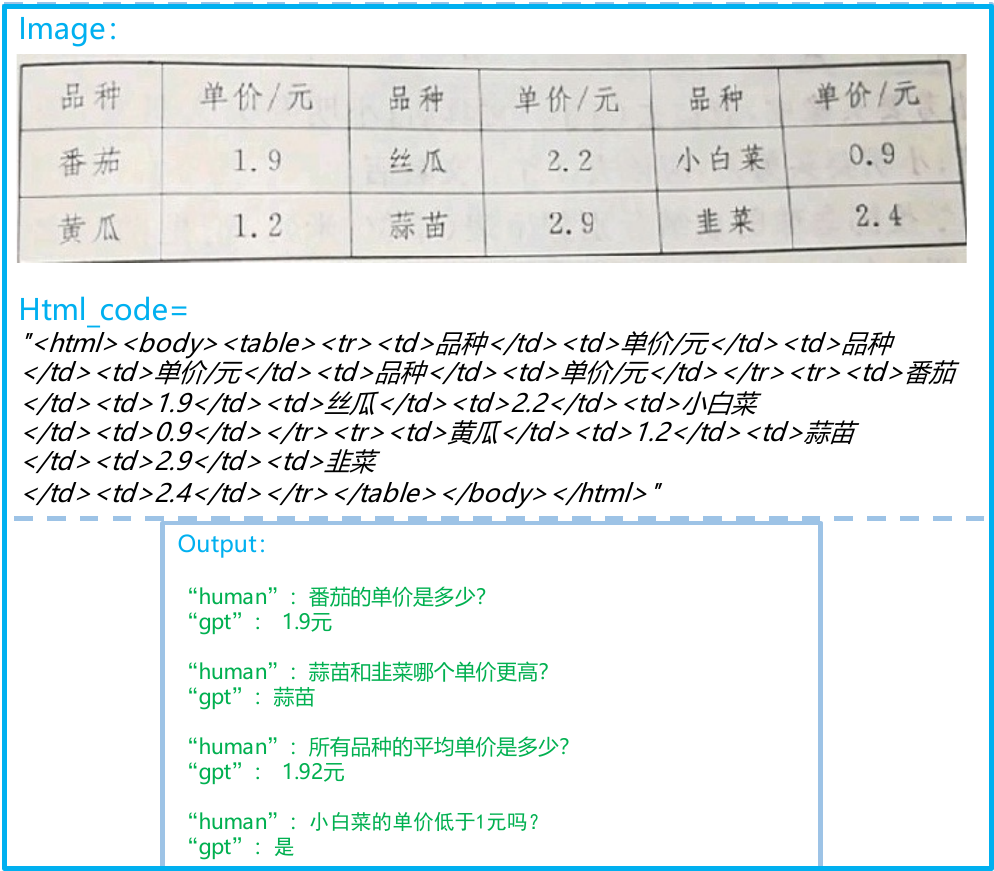}
    \caption{Example for table QA generation}
    \label{fig:table_ex}
\end{figure}

\textbf{Table Images}. Based on the internal table data used for layout analysis tasks (including table data and HTML code), we construct a dataset.\\

\textbf{QA Generation}. We establish a HTML-Table dual-modal alignment mechanism, using the original HTML table structure as baseline information to design hierarchical prompt templates: 1) extract table themes and statistical features for html code, which ensures that the model understands and grasps the values in the table; 2) generate questions that conform to cognitive logic with pre-set task types, which prevents outputting QA that do not conform to human preferences. An example for table QA generation is as Figure~\ref{fig:table_ex}. The prompt used is as follows: \\

\begin{tcolorbox}[colback=white,colframe=black,title=Prompt for Table QA Generation]
question = f\texttt{"}This is a table chart displayed by html code: \{html\_code\}.
Please generate questions of different task types about the chart, and make sure that the answers to the questions can be clearly obtained from the chart, otherwise you can choose not to select the corresponding task.

The selectable task types and their explanations are as follows:

Factoid: Asks a specific fact, the answer comes from a fragment in the table or a value obtained by aggregation.

Free Form: The answer has no fixed format, is usually longer, and is similar to a conversation, such as ChatGPT.

Multiple Choice: The question requires selecting one of multiple options as an answer.

List: The answer requires providing a series of related items, usually from multiple rows or columns in a table.

Yes/No: The answer to the question can only be \texttt{'}yes\texttt{'} or \texttt{'}no\texttt{'}.

Explanation: The answer requires explaining specific data or trends in the table.

Comparison: The answer requires comparing two or more values in the table.

Causal: The answer requires explaining the causal relationship between the data in the table.

Computation: The answer requires some calculations, such as sum, average, etc.

Classification: According to the question, classify or categorize some data in the table.

Time Series: The answer requires analyzing the time series data in the table, such as trends, patterns, etc.

You need to imagine that you are looking at an image rendered by the code, not the code itself. Remember in your answer, all you are given is an image of a graph, and you are answering based on the image.

The values and labels are the ground truth of your question, so make sure the answer is correct.
Avoid using invalid escape characters in strings.
Also, I want to save your output to a json file, so I want you to organize your answer like
\{template\} and must include the answer in \texttt{```json```} format.\texttt{"}
\end{tcolorbox}

\subsubsection{PPInfinityDocData}

\begin{figure}[h]
    \centering
    \includegraphics[width=9cm]{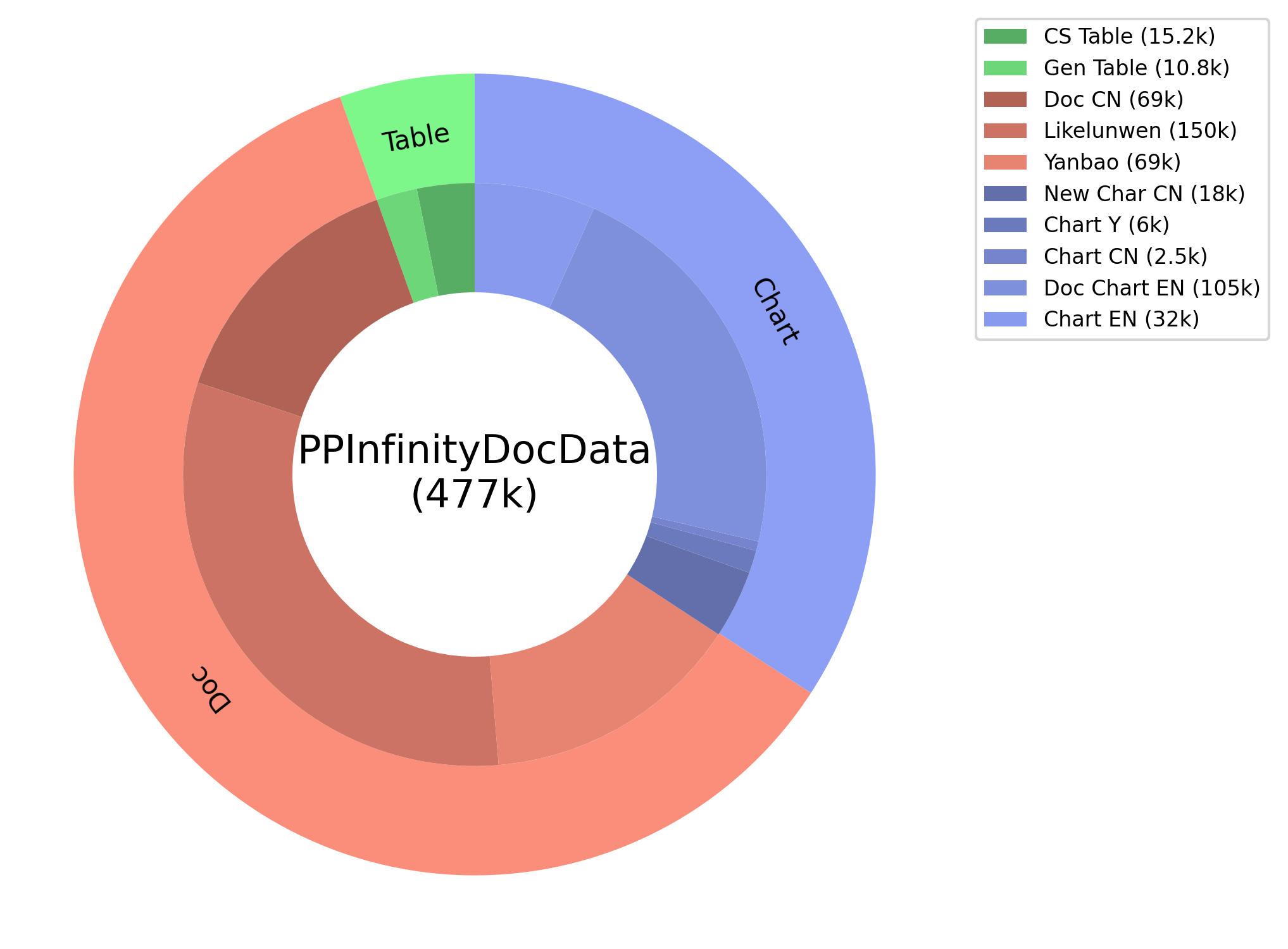}
    \caption{The overall distribution of Synthetic Dataset}
    \label{fig:all_data}
\end{figure}

Owing to the intelligent data generation pipeline, we have successfully generated and screened 477k high-quality Chinese multimodal data, which are collectively named PPInfinityDocData. This dataset has made a significant contribution to the improvement of the final model training. The overall distribution of the synthetic dataset is illustrated in Figure~\ref{fig:all_data}. 

PPInfinityDocData has three categories: Text-rich Document(Doc), Chart and Table. The specific distribution of each category and its subcategories is shown in the figure: the number of Doc is 288k, accounting for 60\% of the total dataset, the number of table is 26k, and the number of Chart is 163k. All Chinese data is 314k, accounting for 66\% of the total dataset.

\section{Methodology}

\subsection{Model Overview}
%
We chose Qwen2-VL~\citep{wang2024qwen2vl, paddlemix2023} as the basic model because of its higher comprehensive accuracy. To deploy quickly and facilitate more low-resource users, we chose the Qwen2-VL-2B model to build our PP-DocBee. 
Although Qwen2-VL-2B has strong document comprehension capabilities, its proficiency is largely confined to English scenarios, given the predominance of English-language training data. In Chinese scenarios, its performance is significantly less developed, indicating substantial potential for improvement.

Applying our "Bag-of-Freebies" technique did not alter the number of parameters or the structure of the basic model. Despite these constraints, it significantly improved the model capability to process English documents and a remarkable enhancement in its ability to handle Chinese documents.

\subsection{Data Pre-processing}

When processing an image, Qwen2-VL employs a patch-based approach, dividing the image into multiple small patches, each typically sized 28×28 pixels, similar to the Vision Transformer (ViT). This method decomposes the image into a series of visual tokens, enabling the model to better understand and process the image content. Notably, the length of Qwen2-VL's visual tokens is dynamic, and the model dynamically adjusts the number of visual tokens based on the resolution of the input image. This adaptive mechanism is particularly effective for Visual Question Answering (VQA) tasks, as it provides more sufficient and comprehensive visual features.

During the training phase, we implemented an expanded range of resize thresholds, increasing the upper limit from 512 pixels to 768 pixels. This approach was designed to augment the overall resolution distribution of the datasets, thereby enriching the visual feature spectrum available to the model. 
In the inference stage, conventional-resolution images were proportionally upscaled by a factor ranging from 1.1 to 1.3. In contrast, images with lower resolutions maintained their original pre-processing strategies. These strategies provided a more comprehensive set of visual features, thereby enhancing the model's understanding capabilities.

\subsection{Dynamic Ratio Sampling Training}

Our training data encompasses diverse document understanding datasets, including general VQA images, OCR images, charts, rich-text documents, mathematical and complex reasoning tasks, synthetic data, and plain text data. We implemented a dynamic data ratio sampling mechanism to optimise the training process and assign different sampling weights to different data and sources. This approach significantly improves the training ratio of high-quality data and balances the quantitative differences between different datasets.

\subsection{OCR Post-process}

Optical Character Recognition (OCR) tools or models are employed to pre-extract text from images through OCR recognition. The extracted text is subsequently provided as auxiliary prior information for the image question. Specifically, the OCR-recognized text is incorporated into the input of the PP-DocBee model during the inference stage by adding a prompt to the original question: "Use the image and the OCR result as context and answer the following question: ". This method effectively enhances model performance, particularly on images containing clear and limited text.

\section{Experiments}

\begin{table*}[t]
	\renewcommand{\arraystretch}{1.5}
	\begin{subtable}{\textwidth}
	    \resizebox{\textwidth}{!}{
		    \begin{tabular}{c|p{17cm}}
			\textbf{Category} & \textbf{Data Composition}\\
			\midrule
			\multirow{2}{*}{\raisebox{+3pt}{Mathematics}} & GeoQA+~\cite{cao2022geoqa_plus}, MathQA~\cite{yu2023mathqa}, CLEVR-Math/Super~\cite{lindstrom2022clevrmath, li2023superclevr}, Geometry3K~\cite{lu2021geometry3k}, MAVIS-math-rule-geo~\cite{zhang2024mavis}, MAVIS-math-metagen~\cite{zhang2024mavis}, GEOS~\cite{seo2015geos}, UniGeo~\cite{chen2022unigeo}\\
                \rowcolor{gray!15}
			\multirow{2}{*}{\raisebox{+3pt}{Science}} & AI2D~\cite{kembhavi2016ai2d}, ScienceQA~\cite{lu2022scienceqa}, TQA~\cite{kembhavi2017tqa}, VisualWebInstruct~\cite{tiger_lab_visualwebinstruct}\\
			\multirow{2}{*}{\raisebox{+3pt}{Chart \& Table}} & ChartQA~\cite{masry2022chartqa}, MMC-Inst~\cite{liu2023mmcinst}, DVQA~\cite{kafle2018dvqa}, PlotQA~\cite{methani2020plotqa}, LRV-Instruction~\cite{liu2023lrv-instruction}, TabMWP~\cite{lu2022tablemwp}, UniChart~\cite{masry2023unichart}, TAT-DQA~\cite{zhu2022tatdqa}, FigureQA~\cite{kahou2017figureqa}, Chart2Text~\cite{kantharaj2022chart2text}, RobuT-\{Wikisql, SQA, WTQ\}~\cite{zhao2023robut}\\
                \rowcolor{gray!15}
			\multirow{5}{*}{\raisebox{+2.75\height}{Naive OCR}} & SynthDoG~\cite{kim2022synthdog}, MTWI~\cite{he2018icpr2018_MTWI}, LVST~\cite{sun2019lsvt}, FUNSD~\cite{jaume2019funsd}, Latex-Formula~\cite{oleehyo_latex_formulas}, IAM~\cite{marti2002iam}, Handwriting-Latex~\cite{aida}, ArT~\cite{chng2019art}, CTW~\cite{yuan2019ctw}, ReCTs~\cite{zhang2019rects}, COCO-Text~\cite{veit2016cocotext}, SVRD~\cite{yu2023icdar_svrd}, MapText~\cite{li2024icdar_maptext}, CAPTCHA~\cite{captcha}, Est-VQA~\cite{wang2020estvqa}, HME-100K~\cite{tal}, TAL-OCR-ENG~\cite{tal}, TAL-HW-MATH~\cite{tal}, IMGUR5K~\cite{krishnan2023textstylebrush_Imgur5K}, Invoices-and-Receipts-OCR~\cite{mychen76_invoices_receipts_ocr_v1}, IIIT5k~\cite{mishra2012scene_iiit5k}, K12-Printing~\cite{tal}, Handwritten-Mathematical-Expression~\cite{Azu}, WordArt~\cite{xie2022toward_wordart}, Handwriting-Forms~\cite{ift_handwriting_forms}\\
			\multirow{4}{*}{\raisebox{+2\height}{OCR QA}} &  DocVQA~\cite{clark2017docqa}, InfoVQA~\cite{mathew2022infographicvqa}, TextVQA~\cite{singh2019textvqa}, ArxivQA~\cite{li2024multimodal_arxivQA}, ScreencQA~\cite{hsiao2022screenqa}, DocReason~\cite{mplug_docreason25k}, Ureader~\cite{ye2023ureader}, FinanceQA~\cite{Sujet-Finance-QA-Vision-100k}, DocMatrix~\cite{laurenccon2024building_docmatrix}, A-OKVQA~\cite{schwenk2022aokvqa}, Diagram-Image-To-Text~\cite{kamizuru00_diagram_image_to_text}, MapQA~\cite{chang2022mapqa}, OCRVQA~\cite{mishra2019ocrvqa}, ST-VQA~\cite{biten2019stvqa}, SQuAD-VQA, VQA-CD~\cite{mahamoud2024chic_vqa_cd}, MTVQA~\cite{tang2024mtvqa}\\
                \rowcolor{gray!15}
			\multirow{5}{*}{\raisebox{+2.75\height}{General VQA}} & LLaVA-150K~\cite{liu2023llava}, LVIS-Instruct4V~\cite{wang2023lvisinstruct4v}, ALLaVA~\cite{chen2024allava},  Laion-GPT4V~\cite{laion_gpt4v_dataset}, LLAVAR~\cite{zhang2023llavar}, VizWiz~\cite{gurari2018vizwiz}, MMinstruct~\cite{liu2024mminstruct}, WildVision~\cite{lu2024wildvision}, LLaVA-Critic-113k~\cite{xiong2024llava_critic}, VQAv2~\cite{goyal2017vqav2}, MMRA~\cite{wu2024mmra}, MMDU~\cite{liu2024mmdu}, IconQA~\cite{lu2021iconqa}\\
			\multirow{2}{*}{\raisebox{+1.3\height}{Text-only}} & WizardLM~\cite{xu2023wizardlm}, Infinity-Instruct~\cite{baai_infinity_instruct}, UltraInteract-sft~\cite{yuan2024advancing_ultrainteract}\\
			\end{tabular}
		}
		\vspace{-1mm}
		\caption{Summary of the open source public dataset used in PP-DocBee.}
		\label{tab:dataset_open}
		\vspace{-1mm}
	\end{subtable}
	
        \vspace{10pt}
    
	\begin{subtable}{\textwidth}
		\resizebox{\textwidth}{!}{
		\begin{tabular}{c|c|p{17cm}}
			\textbf{Category} & \textbf{Total number of samples} & \textbf{Data Composition}\\
			\midrule
			Text-rich Document & 288k & Financial Reports, Research Reports, Laws and Regulations, Science and Engineering Papers\\
			\rowcolor{gray!15}
			Table & 26k & Instructions, Science and Engineering Papers\\
			Chart & 163k & Contracts, Papers\\
		\end{tabular}}
		\vspace{-1mm}
		\caption{Summary of the synthetic dataset PPInfinityDocData used in PP-DocBee.}
		\label{tab:dataset_sys}
		\vspace{-1mm}
	\end{subtable}
	\caption{SFT Dataset used in PP-DocBee.}
	\label{tab:dataset_all}
\end{table*}

\subsection{Training Data}
The SFT datasets used in PP-DocBee are detailed in Table~\ref{tab:dataset_all}. This comprehensive collection includes a diverse array of public datasets, as described in Table~\ref{tab:dataset_open}, as well as internally generated synthetic datasets, which are described in Table~\ref{tab:dataset_sys}. Integrating these datasets ensures a rich and varied training environment and improves the generalization ability across different scenarios and modalities.

\subsection{Implementation Details}
PP-DocBee is initialized from the Qwen2-VL-2B model~\citep{wang2024qwen2vl}, which employs a Vision Transformer (ViT) with approximately 675 million parameters as the visual encoder and a 1.5B Qwen2 Large Language Model (LLM) as the language decoder. 
During the supervised fine-tuning (SFT) stage, the visual encoder is frozen while the parameters of the LLM are updated. PP-DocBee is trained for 16k iterations on a dataset comprising nearly 5 million samples, with a batch size of 32. This training process takes approximately 2 days using single node with 8 NVIDIA A800 GPUs.

\subsection{Comparison with SOTA}
We evaluated the performance of PP-DocBee on five English text-rich image benchmarks and our internal Chinese business scenario image benchmarks. PP-DocBee was compared with existing state-of-the-art OCR-free multimodal large language models (MLLMs) of the same parameter size, as well as closed-source APIs.
The five English benchmarks cover documents (DocVQA~\citep{docvqa}, InfoVQA~\citep{infovqa}), charts (ChartQA~\citep{chartqa}), natural images (TextVQA~\citep{textvqa}), and OCR-related images (OCRBench~\citep{liu2023ocrbench}). Our internal Chinese business evaluation set includes financial reports, laws and regulations, scientific and engineering papers, instructions, liberal arts papers, contracts, research reports, and other relevant scenes. The resolution of all images is very high, with an average resolution of approximately 1680×1204, comprising a total of 1196 data samples. These can be categorized into four types: printed text (656 images), tables (358 images), seals (15 images), and charts (167 images).

\begin{table*}
  \centering
  \renewcommand{\arraystretch}{0.95} 
  \addtolength{\tabcolsep}{2pt} 
  \fontsize{9}{12}\selectfont
  \begin{tabular}{ 
                  @{}
                  l
                  >{\centering\arraybackslash}p{1cm}
                  >{\centering\arraybackslash}p{1cm}
                  >{\centering\arraybackslash}p{1cm}
                  >{\centering\arraybackslash}p{1cm}
                  >{\centering\arraybackslash}p{1cm}
                  @{}
                  }
    \toprule
    Method & 
    \multicolumn{1}{>{\centering\arraybackslash}p{1cm}}{\fontsize{8}{9}\selectfont DocVQA-test} & 
    \multicolumn{1}{>{\centering\arraybackslash}p{1cm}}{\fontsize{8}{9}\selectfont ChartQA} & 
    \multicolumn{1}{>{\centering\arraybackslash}p{1cm}}{\fontsize{8}{9}\selectfont InfoVQA-test} & 
    \multicolumn{1}{>{\centering\arraybackslash}p{1cm}}{\fontsize{8}{9}\selectfont TextVQA} & 
    \multicolumn{1}{>{\centering\arraybackslash}p{1cm}}{\fontsize{8}{9}\selectfont OCRBench} \\
    \midrule
    \rowcolor{gray!20}
    \multicolumn{6}{c}{Closed-source MLMMs} \\
    GPT-4o~\cite{openai2024gpt4o} & 92.8 & 85.7 & 79.2 & 77.4 & 73.6 \\
    Gemini-1.5-Pro~\cite{team2023gemini} & 93.1 & 87.2 & 80.1 & 78.7 & 75.4 \\
    \rowcolor{gray!20}
    \multicolumn{6}{c}{Open-source MLMMs} \\
    MiniCPM-V-2-2B~\cite{yao2024minicpm} & 71.9 & - & - & 74.1 & 60.5 \\
    Aquila-VL-2B~\cite{gu2024infinitymmscalingmultimodalperformance} & 85.0 & 76.5 & 58.3 & 76.4 & 77.2 \\
    Mini-Monkey-2B~\cite{huang2024mini} & 87.4 & 76.5 & 60.1 & 76.0 & 79.4 \\
    InternVL2-2B~\cite{chen2024internvl} & 86.9 & 76.2 & 58.9 & 73.4 & 78.1 \\
    InternVL2.5-2B~\cite{chen2024expanding} & 88.7 & \textbf{79.2} & 60.9 & 74.3 & 80.4 \\
    Qwen2-VL-2B~\cite{wang2024qwen2vl} & 90.1 & 73.5 & 65.5 & 79.7 & 80.9(82.2) \\
    \textbf{PP-DocBee-2B} & \textbf{90.6} & 74.6 & \textbf{66.2} & \textbf{81.2} & \textbf{82.8(83.5)} \\
    \bottomrule
  \end{tabular}
  \vspace{-5pt}
  \caption{Evaluation of existing OCR-Free MLLMs on public benchmarks. The 83.5 in brackets after 82.8 means the score after using OCR post-processing assistance.}
  \label{tab:English_subsets}
  \vspace{-5pt}
\end{table*}

As shown in Table~\ref{tab:English_subsets}, We compare PP-DocBee with the existing state-of-the-art OCR-free MLLMs and APIs.
The table lists the performance indicators of multiple open-source and closed-source models on five tasks: DocVQA, ChantQA, InfoVQA, TextVQA, and OCRBench. 
By comparing these indicators, we found that the PP-DocBee-2B model showed excellent performance on multiple tasks, especially on the TextVQA task, which achieved a high score of 81.2, and on the OCRBench task, it also achieved a high score of 82.8; both of which are the highest scores among all models in the table. After using OCR post-processing assistance in the OCRBench task, we got a higher score of 83.5. This strategy also proved effective on Qwen2-VL, as shown in the table, from 80.9 points to 82.2 points. We found this strategy effectively enhances images containing clear and limited text.
Considering the performance of all tasks, the PP-DocBee-2B model performed the best with comprehensive accuracy, which shows that it has high accuracy and reliability when dealing with Chinese-related multimodal tasks.

Table~\ref{tab:Chinese_subsets} illustrates the performance of the PP-DocBee-2B model across multiple categories. Notably, in the``Painted text" category, the PP-DocBee-2B model achieved a leading score of 517, outperforming other models.
Furthermore, the PP-DocBee-2B model achieved a high score of 202 in the ``Tables" category, demonstrating its proficiency in understanding and processing tabular data. Although the scores in the``Seals" and``Charts" categories were slightly lower, at 5 and 41, respectively, the overall score of 765 was the highest among all models. 
This result indicates that PP-DocBee-2B exhibits a high comprehensive accuracy when processing Chinese multimodal data, and it has demonstrated strong capabilities in text recognition, table parsing, and understanding other visual elements. Therefore, it can be concluded that PP-DocBee-2B achieved the best performance in terms of comprehensive accuracy. A more comprehensive and high-quality presentation of some cases can be found in Appendix A.

\begin{table}
  \centering
  \renewcommand{\arraystretch}{0.95} 
  \addtolength{\tabcolsep}{2pt} 
  \fontsize{9}{12}\selectfont
  \begin{tabular}{ 
                  @{}
                  l
                  >{\centering\arraybackslash}p{1cm}
                  >{\centering\arraybackslash}p{1cm}
                  >{\centering\arraybackslash}p{1cm}
                  >{\centering\arraybackslash}p{1cm}
                  >{\centering\arraybackslash}p{1cm}
                  @{}
                  }
    \toprule
    Method & 
    \multicolumn{1}{>{\centering\arraybackslash}p{1cm}}{\fontsize{8}{9}\selectfont Printed text} & 
    \multicolumn{1}{>{\centering\arraybackslash}p{1cm}}{\fontsize{8}{9}\selectfont Tables} & 
    \multicolumn{1}{>{\centering\arraybackslash}p{1cm}}{\fontsize{8}{9}\selectfont Seals} & 
    \multicolumn{1}{>{\centering\arraybackslash}p{1cm}}{\fontsize{8}{9}\selectfont Charts} & 
    \multicolumn{1}{>{\centering\arraybackslash}p{1cm}}{\fontsize{8}{9}\selectfont Total Score} \\
    \midrule
    \rowcolor{gray!20}
    \multicolumn{6}{c}{Closed-source MLMMs} \\
    GPT-4o~\cite{openai2024gpt4o} & 436 & 198 & 5 & 46 & 685 \\
    GLM-4V Flash~\cite{glm2024chatglm} & 339 & 169 & 5 & 34 & 547 \\
    \rowcolor{gray!20}
    \multicolumn{6}{c}{Open-source MLMMs} \\
    InternVL2.5-2B~\cite{chen2024expanding} & 363 & 182 & 4 & 47 & 596 \\
    Qwen2-VL-2B~\cite{wang2024qwen2vl} & 476 & 167 & 8 & 29 & 680 \\
    \textbf{PP-DocBee-2B} & \textbf{517} & \textbf{202} & 5 & 41 & \textbf{765} \\
    \bottomrule
  \end{tabular}
  \vspace{-5pt}
  \caption{Evaluation of existing OCR-Free MLLMs on internal Chinese benchmarks.}
  \label{tab:Chinese_subsets}
  \vspace{-5pt}
\end{table}

\begin{table*}
  \centering
  \renewcommand{\arraystretch}{0.95} 
  \addtolength{\tabcolsep}{1.5pt} 
  \fontsize{9}{12}\selectfont
  \begin{tabular}{ 
                  @{}
                  l
                  >{\centering\arraybackslash}p{1cm}
                  >{\centering\arraybackslash}p{1cm}
                  >{\centering\arraybackslash}p{1cm}
                  >{\centering\arraybackslash}p{1cm}
                  >{\centering\arraybackslash}p{1cm}
                  >{\centering\arraybackslash}p{1cm}
                  @{}
                  }
    \toprule
    Method & 
    \multicolumn{1}{>{\centering\arraybackslash}p{1cm}}{\fontsize{8}{9}\selectfont DocVQA-val} & 
    \multicolumn{1}{>{\centering\arraybackslash}p{1cm}}{\fontsize{8}{9}\selectfont ChartQA} & 
    \multicolumn{1}{>{\centering\arraybackslash}p{1cm}}{\fontsize{8}{9}\selectfont InfoVQA-val} & 
    \multicolumn{1}{>{\centering\arraybackslash}p{1cm}}{\fontsize{8}{9}\selectfont TextVQA} & 
    \multicolumn{1}{>{\centering\arraybackslash}p{1cm}}{\fontsize{8}{9}\selectfont OCRBench} &
    \multicolumn{1}{>{\centering\arraybackslash}p{1cm}}{\fontsize{8}{9}\selectfont Internal-CN} \\
    \midrule
    Baseline            & 89.2 & 73.5 & 64.1 & 79.7 & 80.9 & 680 \\
    Baseline + 3.3M     & 89.6 & 74.3 & 65.0 & 80.6 & 81.6 & 726 \\
    Baseline + PPIDD  & 89.3 & 73.7 &64.2 & 79.9 & 81.0 & 725 \\
    Baseline + 3.3M + PPIDD(12\%) & 89.6 & 74.6 & 65.0 & 80.6 & 81.6 & 743 \\
    Baseline + 3.3M + PPIDD(12\%) + MS & 89.7 & 74.6 & 65.0 & 80.6 & 81.4 & 745 \\
    Baseline + 3.3M + PPIDD(20\%) + MS & 89.7 & 74.6 & 65.2 & 80.6 & 80.7 & 765 \\
    Baseline + 3.3M + PPIDD(30\%) + MS & 89.7 & 74.4 & 64.8 & 80.6 & 80.9 & 761 \\
    \textbf{PP-DocBee-2B} & \textbf{90.1} & 74.6 & \textbf{65.4} & \textbf{81.2} & \textbf{82.8} & \textbf{765} \\
    \bottomrule
  \end{tabular}
  \vspace{-5pt}
  \caption{Ablation study of PP-DocBee.``3.3M" means adding 3.3M open source public datasets.``PPIDD" means adding 477k synthetic dataset PPInfinityDocData. The number in the brackets after it, such as (20\%), indicates the proportion of synthetic data to the total training data. If 3.3M public datasets and 477k PPInfinityDocData are added, the default proportion of synthetic data is 12\%. This ratio adjustment represents the use of the Dynamic Ratio Sampling Training strategy.``MS" means more steps, the default is 16k steps. Based on``Baseline + 3.3M + PPIDD + MS", PP-DocBee in the last row of the table also uses data preprocessing, which is to enlarge the image input scale.}
  \label{tab:ablation_study}
  \vspace{-5pt}
\end{table*}

\subsection{Ablation Study}

To establish baseline performance, we initialized the Qwen2-VL-2B-Instruct model with pre-trained weights and extended its training using 330k samples from dvqa, chartqa, ai2d, docvqa, geoqa+ and synthdog. 
For comprehensive evaluation, we use five established English benchmark datasets as well as our proprietary internal Chinese scene annotation evaluation dataset (named Internal-CN).
Table~\ref{tab:ablation_study} presents a systematic ablation study evaluating the efficacy of two key innovations:  Data Synthesis Strategy and Dynamic Ratio Sampling.

\noindent\textbf{Effectiveness of Data Synthesis Strategy.} To assess the impact of data synthesis strategies, we conducted a series of ablation experiments. These experiments compare the performance of the baseline model with the performance of models trained in different settings.
We can see that on Internal-CN, the setting of``Baseline + PPIDD" (adding 477k synthetic data PPInfinityDocData to the baseline) can reach 725 points, while the setting of``Baseline + 3.3M" (adding 3.3M open source data to the baseline) can also reach 726 points. Because our Internal-CN evaluation set is closer to the real application scenario, it is closer to the source of the synthetic dataset, and has some dominance with the open source dataset. However, adding only 477k synthetic data has limited gains for five established English benchmark datasets.
These results show that adding synthetic data (especially when tailored to specific weaknesses in the real dataset) can significantly improve the performance of the model's Chinese document scene understanding.

\noindent\textbf{Effectiveness of Dynamic Ratio Sampling.} We also conducted experiments to evaluate the effect of dynamic ratio sampling. After adding the 3.3M open source dataset and the 477k synthetic dataset, our training dataset contains English and Chinese document images from different fields, including financial reports, scientific articles, and legal documents. As can be seen in the overall table, the same setting of ``Baseline + 3.3M + PPIDD + MS" is used, but the ratio of 20\% is the best, with a maximum score of 765. Even if the ratio is increased to 30\%, the score is only 761. This experimental result shows that assigning appropriate sampling weights to the training dataset according to the representativeness of the data and the characteristics of the evaluation set can significantly improve the ability of the model in some aspects.

\section{Conclusion}
\label{sec:conclusion}

PP-DocBee has made progress in the field of document image understanding through two major innovations: (1) a data synthesis strategy for Chinese document scenarios; (2) a set of training tricks, such as dynamic proportional sampling training, data preprocessing, OCR post-processing, etc. These innovations have greatly improved the model's ability to process and understand complex document images, setting a new benchmark in this field.

In future, we plan to conduct extensive experiments on a broader range of models and delve into more specialized areas, such as handwritten document analysis. We anticipate that the comprehensive solution offered by PP-DocBee will be well-equipped to tackle intricate document analysis tasks, thereby propelling the state-of-the-art in document image understanding.

\clearpage
\bibliography{iclr2025_conference}

\begin{thebibliography}{105}
\providecommand{\natexlab}[1]{#1}
\providecommand{\url}[1]{\texttt{#1}}
\expandafter\ifx\csname urlstyle\endcsname\relax
  \providecommand{\doi}[1]{doi: #1}\else
  \providecommand{\doi}{doi: \begingroup \urlstyle{rm}\Url}\fi

\bibitem[aidapearson(2023)]{aida}
aidapearson.
\newblock Aida calculus math handwriting recognition dataset.
\newblock \url{https://www.kaggle.com/datasets/aidapearson/ocr-data}, 2023.

\bibitem[Authors(2023)]{paddlemix2023}
PaddlePaddle Authors.
\newblock Paddlemix, paddle multimodal integration and exploration.
\newblock \url{https://github.com/PaddlePaddle/PaddleMIX}, 2023.

\bibitem[Azu(2023)]{Azu}
Azu.
\newblock Handwritten-mathematical-expression-convert-latex.
\newblock \url{https://huggingface.co/datasets/Azu/Handwritten-Mathematical-Expression-Convert-LaTeX}, 2023.

\bibitem[BAAI(2024)]{baai_infinity_instruct}
BAAI.
\newblock Infinity-instruct dataset.
\newblock \url{https://huggingface.co/datasets/BAAI/Infinity-Instruct}, 2024.

\bibitem[Bai et~al.(2023{\natexlab{a}})Bai, Bai, Chu, Cui, Dang, Deng, Fan, Ge, Han, Huang, Hui, Ji, Li, Lin, Lin, Liu, Liu, Lu, Lu, Ma, Men, Ren, Ren, Tan, Tan, Tu, Wang, Wang, Wang, Wu, Xu, Xu, Yang, Yang, Yang, Yang, Yao, Yu, Yuan, Yuan, Zhang, Zhang, Zhang, Zhang, Zhou, Zhou, Zhou, and Zhu]{qwen}
Jinze Bai, Shuai Bai, Yunfei Chu, Zeyu Cui, Kai Dang, Xiaodong Deng, Yang Fan, Wenbin Ge, Yu~Han, Fei Huang, Binyuan Hui, Luo Ji, Mei Li, Junyang Lin, Runji Lin, Dayiheng Liu, Gao Liu, Chengqiang Lu, Keming Lu, Jianxin Ma, Rui Men, Xingzhang Ren, Xuancheng Ren, Chuanqi Tan, Sinan Tan, Jianhong Tu, Peng Wang, Shijie Wang, Wei Wang, Shengguang Wu, Benfeng Xu, Jin Xu, An~Yang, Hao Yang, Jian Yang, Shusheng Yang, Yang Yao, Bowen Yu, Hongyi Yuan, Zheng Yuan, Jianwei Zhang, Xingxuan Zhang, Yichang Zhang, Zhenru Zhang, Chang Zhou, Jingren Zhou, Xiaohuan Zhou, and Tianhang Zhu.
\newblock Qwen technical report.
\newblock \emph{arXiv preprint arXiv:2309.16609}, 2023{\natexlab{a}}.

\bibitem[Bai et~al.(2023{\natexlab{b}})Bai, Bai, Yang, Wang, Tan, Wang, Lin, Zhou, and Zhou]{qwenvl}
Jinze Bai, Shuai Bai, Shusheng Yang, Shijie Wang, Sinan Tan, Peng Wang, Junyang Lin, Chang Zhou, and Jingren Zhou.
\newblock Qwen-vl: A versatile vision-language model for understanding, localization, text reading, and beyond.
\newblock \emph{arXiv preprint arXiv:2308.12966}, 2023{\natexlab{b}}.

\bibitem[Biten et~al.(2019)Biten, Tito, Mafla, Gomez, Rusinol, Valveny, Jawahar, and Karatzas]{biten2019stvqa}
Ali~Furkan Biten, Ruben Tito, Andres Mafla, Lluis Gomez, Mar{\c{c}}al Rusinol, Ernest Valveny, CV~Jawahar, and Dimosthenis Karatzas.
\newblock Scene text visual question answering.
\newblock In \emph{ICCV}, pp.\  4291--4301, 2019.

\bibitem[Cao \& Xiao(2022)Cao and Xiao]{cao2022geoqa_plus}
Jie Cao and Jing Xiao.
\newblock An augmented benchmark dataset for geometric question answering through dual parallel text encoding.
\newblock In \emph{Proceedings of the 29th International Conference on Computational Linguistics}, pp.\  1511--1520, 2022.

\bibitem[Chang et~al.(2022)Chang, Palzer, Li, Fosler-Lussier, and Xiao]{chang2022mapqa}
Shuaichen Chang, David Palzer, Jialin Li, Eric Fosler-Lussier, and Ningchuan Xiao.
\newblock Mapqa: A dataset for question answering on choropleth maps.
\newblock \emph{arXiv:2211.08545}, 2022.

\bibitem[Chen et~al.(2024{\natexlab{a}})Chen, Chen, Zhang, Chen, Wu, Zhang, Chen, Li, Wan, and Wang]{chen2024allava}
Guiming~Hardy Chen, Shunian Chen, Ruifei Zhang, Junying Chen, Xiangbo Wu, Zhiyi Zhang, Zhihong Chen, Jianquan Li, Xiang Wan, and Benyou Wang.
\newblock Allava: Harnessing gpt4v-synthesized data for a lite vision-language model.
\newblock \emph{arXiv preprint arXiv:2402.11684}, 2024{\natexlab{a}}.

\bibitem[Chen et~al.(2022)Chen, Li, Qin, Lu, Lin, Chen, and Liang]{chen2022unigeo}
Jiaqi Chen, Tong Li, Jinghui Qin, Pan Lu, Liang Lin, Chongyu Chen, and Xiaodan Liang.
\newblock Unigeo: Unifying geometry logical reasoning via reformulating mathematical expression.
\newblock \emph{arXiv preprint arXiv:2212.02746}, 2022.

\bibitem[Chen et~al.(2024{\natexlab{b}})Chen, Wang, Cao, Liu, Gao, Cui, Zhu, Ye, Tian, Liu, et~al.]{chen2024expanding}
Zhe Chen, Weiyun Wang, Yue Cao, Yangzhou Liu, Zhangwei Gao, Erfei Cui, Jinguo Zhu, Shenglong Ye, Hao Tian, Zhaoyang Liu, et~al.
\newblock Expanding performance boundaries of open-source multimodal models with model, data, and test-time scaling.
\newblock \emph{arXiv preprint arXiv:2412.05271}, 2024{\natexlab{b}}.

\bibitem[Chen et~al.(2024{\natexlab{c}})Chen, Wu, Wang, Su, Chen, Xing, Zhong, Zhang, Zhu, Lu, et~al.]{chen2024internvl}
Zhe Chen, Jiannan Wu, Wenhai Wang, Weijie Su, Guo Chen, Sen Xing, Muyan Zhong, Qinglong Zhang, Xizhou Zhu, Lewei Lu, et~al.
\newblock Internvl: Scaling up vision foundation models and aligning for generic visual-linguistic tasks.
\newblock In \emph{Proceedings of the IEEE/CVF Conference on Computer Vision and Pattern Recognition}, pp.\  24185--24198, 2024{\natexlab{c}}.

\bibitem[Cheng et~al.(2023)Cheng, Zhang, Wu, Zhang, Zhu, Xie, Li, Ding, and Jin]{Cheng_2023_CVPR}
Hiuyi Cheng, Peirong Zhang, Sihang Wu, Jiaxin Zhang, Qiyuan Zhu, Zecheng Xie, Jing Li, Kai Ding, and Lianwen Jin.
\newblock M6doc: A large-scale multi-format, multi-type, multi-layout, multi-language, multi-annotation category dataset for modern document layout analysis.
\newblock In \emph{Proceedings of the IEEE/CVF Conference on Computer Vision and Pattern Recognition (CVPR)}, pp.\  15138--15147, June 2023.

\bibitem[Chng et~al.(2019)Chng, Liu, Sun, Ng, Luo, Ni, Fang, Zhang, Han, Ding, et~al.]{chng2019art}
Chee~Kheng Chng, Yuliang Liu, Yipeng Sun, Chun~Chet Ng, Canjie Luo, Zihan Ni, ChuanMing Fang, Shuaitao Zhang, Junyu Han, Errui Ding, et~al.
\newblock Icdar2019 robust reading challenge on arbitrary-shaped text-rrc-art.
\newblock pp.\  1571--1576, 2019.

\bibitem[Clark \& Gardner(2018)Clark and Gardner]{clark2017docqa}
Christopher Clark and Matt Gardner.
\newblock Simple and effective multi-paragraph reading comprehension.
\newblock pp.\  845--855, 2018.

\bibitem[Dosovitskiy et~al.(2021)Dosovitskiy, Beyer, Kolesnikov, Weissenborn, Zhai, Unterthiner, Dehghani, Minderer, Heigold, Gelly, Uszkoreit, and Houlsby]{vit2021}
Alexey Dosovitskiy, Lucas Beyer, Alexander Kolesnikov, Dirk Weissenborn, Xiaohua Zhai, Thomas Unterthiner, Mostafa Dehghani, Matthias Minderer, Georg Heigold, Sylvain Gelly, Jakob Uszkoreit, and Neil Houlsby.
\newblock An image is worth 16x16 words: Transformers for image recognition at scale.
\newblock In \emph{{ICLR}}. OpenReview.net, 2021.

\bibitem[GLM et~al.(2024)GLM, Zeng, Xu, Wang, Zhang, Yin, Rojas, Feng, Zhao, Lai, et~al.]{glm2024chatglm}
Team GLM, Aohan Zeng, Bin Xu, Bowen Wang, Chenhui Zhang, Da~Yin, Diego Rojas, Guanyu Feng, Hanlin Zhao, Hanyu Lai, et~al.
\newblock {ChatGLM: A Family of Large Language Models from GLM-130B to GLM-4 All Tools}.
\newblock \emph{arXiv preprint arXiv:2406.12793}, 2024.

\bibitem[Goyal et~al.(2017)Goyal, Khot, Summers-Stay, Batra, and Parikh]{goyal2017vqav2}
Yash Goyal, Tejas Khot, Douglas Summers-Stay, Dhruv Batra, and Devi Parikh.
\newblock Making the v in vqa matter: Elevating the role of image understanding in visual question answering.
\newblock In \emph{CVPR}, pp.\  6904--6913, 2017.

\bibitem[Gu et~al.(2024)Gu, Zhang, Zhou, Yu, Xing, Wang, Cao, Jia, Zhang, Wang, Hu, Zhang, Li, Liang, Zhao, Ao, Liu, Feng, and Liu]{gu2024infinitymmscalingmultimodalperformance}
Shuhao Gu, Jialing Zhang, Siyuan Zhou, Kevin Yu, Zhaohu Xing, Liangdong Wang, Zhou Cao, Jintao Jia, Zhuoyi Zhang, Yixuan Wang, Zhenchong Hu, Bo-Wen Zhang, Jijie Li, Dong Liang, Yingli Zhao, Yulong Ao, Yaoqi Liu, Fangxiang Feng, and Guang Liu.
\newblock Infinity-mm: Scaling multimodal performance with large-scale and high-quality instruction data, 2024.
\newblock URL \url{https://arxiv.org/abs/2410.18558}.

\bibitem[Gurari et~al.(2018)Gurari, Li, Stangl, Guo, Lin, Grauman, Luo, and Bigham]{gurari2018vizwiz}
Danna Gurari, Qing Li, Abigale~J Stangl, Anhong Guo, Chi Lin, Kristen Grauman, Jiebo Luo, and Jeffrey~P Bigham.
\newblock Vizwiz grand challenge: Answering visual questions from blind people.
\newblock In \emph{CVPR}, pp.\  3608--3617, 2018.

\bibitem[He et~al.(2018)He, Liu, Yang, Zhang, Luo, Gao, Zheng, Wang, Zhang, and Jin]{he2018icpr2018_MTWI}
Mengchao He, Yuliang Liu, Zhibo Yang, Sheng Zhang, Canjie Luo, Feiyu Gao, Qi~Zheng, Yongpan Wang, Xin Zhang, and Lianwen Jin.
\newblock Icpr2018 contest on robust reading for multi-type web images.
\newblock In \emph{ICPR}, 2018.

\bibitem[Hsiao et~al.(2022)Hsiao, Zubach, Baechler, Carbune, Lin, Wang, Sunkara, Zhu, and Chen]{hsiao2022screenqa}
Yu-Chung Hsiao, Fedir Zubach, Gilles Baechler, Victor Carbune, Jason Lin, Maria Wang, Srinivas Sunkara, Yun Zhu, and Jindong Chen.
\newblock Screenqa: Large-scale question-answer pairs over mobile app screenshots.
\newblock \emph{arXiv preprint arXiv:2209.08199}, 2022.

\bibitem[Hu et~al.(2024)Hu, Xu, Ye, Yan, Zhang, Zhang, Li, Zhang, Jin, Huang, et~al.]{mplug_docreason25k}
Anwen Hu, Haiyang Xu, Jiabo Ye, Ming Yan, Liang Zhang, Bo~Zhang, Chen Li, Ji~Zhang, Qin Jin, Fei Huang, et~al.
\newblock mplug-docowl 1.5: Unified structure learning for ocr-free document understanding.
\newblock \emph{arXiv:2403.12895}, 2024.

\bibitem[Huang et~al.(2024)Huang, Liu, Liang, Jin, and Bai]{huang2024mini}
Mingxin Huang, Yuliang Liu, Dingkang Liang, Lianwen Jin, and Xiang Bai.
\newblock Mini-monkey: Multi-scale adaptive cropping for multimodal large language models.
\newblock \emph{arXiv preprint arXiv:2408.02034}, 2024.

\bibitem[ift(2024)]{ift_handwriting_forms}
ift.
\newblock Handwriting forms dataset.
\newblock \url{https://huggingface.co/datasets/ift/handwriting_forms}, 2024.

\bibitem[Jaume et~al.(2019)Jaume, Ekenel, and Thiran]{jaume2019funsd}
Guillaume Jaume, Hazim~Kemal Ekenel, and Jean-Philippe Thiran.
\newblock Funsd: A dataset for form understanding in noisy scanned documents.
\newblock In \emph{ICDAR Workshops}, 2019.

\bibitem[Kafle et~al.(2018)Kafle, Price, Cohen, and Kanan]{kafle2018dvqa}
Kushal Kafle, Brian Price, Scott Cohen, and Christopher Kanan.
\newblock Dvqa: Understanding data visualizations via question answering.
\newblock In \emph{CVPR}, pp.\  5648--5656, 2018.

\bibitem[Kahou et~al.(2017)Kahou, Michalski, Atkinson, K{\'a}d{\'a}r, Trischler, and Bengio]{kahou2017figureqa}
Samira~Ebrahimi Kahou, Vincent Michalski, Adam Atkinson, {\'A}kos K{\'a}d{\'a}r, Adam Trischler, and Yoshua Bengio.
\newblock Figureqa: An annotated figure dataset for visual reasoning.
\newblock \emph{arXiv preprint arXiv:1710.07300}, 2017.

\bibitem[Kamizuru00(2024)]{kamizuru00_diagram_image_to_text}
Kamizuru00.
\newblock Diagram image to text dataset.
\newblock \url{https://huggingface.co/datasets/Kamizuru00/diagram_image_to_text}, 2024.

\bibitem[Kantharaj et~al.(2022)Kantharaj, Leong, Lin, Masry, Thakkar, Hoque, and Joty]{kantharaj2022chart2text}
Shankar Kantharaj, Rixie Tiffany~Ko Leong, Xiang Lin, Ahmed Masry, Megh Thakkar, Enamul Hoque, and Shafiq Joty.
\newblock Chart-to-text: A large-scale benchmark for chart summarization.
\newblock \emph{arXiv:2203.06486}, 2022.

\bibitem[Kembhavi et~al.(2016)Kembhavi, Salvato, Kolve, Seo, Hajishirzi, and Farhadi]{kembhavi2016ai2d}
Aniruddha Kembhavi, Mike Salvato, Eric Kolve, Minjoon Seo, Hannaneh Hajishirzi, and Ali Farhadi.
\newblock A diagram is worth a dozen images.
\newblock In \emph{ECCV}, pp.\  235--251, 2016.

\bibitem[Kembhavi et~al.(2017)Kembhavi, Seo, Schwenk, Choi, Farhadi, and Hajishirzi]{kembhavi2017tqa}
Aniruddha Kembhavi, Minjoon Seo, Dustin Schwenk, Jonghyun Choi, Ali Farhadi, and Hannaneh Hajishirzi.
\newblock Are you smarter than a sixth grader? textbook question answering for multimodal machine comprehension.
\newblock In \emph{CVPR}, pp.\  4999--5007, 2017.

\bibitem[Kim et~al.(2022)Kim, Hong, Yim, Nam, Park, Yim, Hwang, Yun, Han, and Park]{kim2022synthdog}
Geewook Kim, Teakgyu Hong, Moonbin Yim, JeongYeon Nam, Jinyoung Park, Jinyeong Yim, Wonseok Hwang, Sangdoo Yun, Dongyoon Han, and Seunghyun Park.
\newblock {OCR-Free} document understanding transformer.
\newblock In \emph{ECCV}, 2022.

\bibitem[Krishnan et~al.(2023)Krishnan, Kovvuri, Pang, Vassilev, and Hassner]{krishnan2023textstylebrush_Imgur5K}
Praveen Krishnan, Rama Kovvuri, Guan Pang, Boris Vassilev, and Tal Hassner.
\newblock Textstylebrush: transfer of text aesthetics from a single example.
\newblock \emph{IEEE Trans. PAMI}, 45\penalty0 (7):\penalty0 9122--9134, 2023.

\bibitem[LAION(2023)]{laion_gpt4v_dataset}
LAION.
\newblock Gpt-4v dataset.
\newblock \url{https://huggingface.co/datasets/laion/gpt4v-dataset}, 2023.

\bibitem[Lauren{\c{c}}on et~al.(2024)Lauren{\c{c}}on, Marafioti, Sanh, and Tronchon]{laurenccon2024building_docmatrix}
Hugo Lauren{\c{c}}on, Andr{\'e}s Marafioti, Victor Sanh, and L{\'e}o Tronchon.
\newblock Building and better understanding vision-language models: insights and future directions.
\newblock \emph{arXiv:2408.12637}, 2024.

\bibitem[Li et~al.(2022)Li, Liu, Guo, Yin, Jiang, Du, Du, Zhu, Lai, Hu, et~al.]{li2022paddleocr}
Chenxia Li, Weiwei Liu, Ruoyu Guo, Xiaoting Yin, Kaitao Jiang, Yongkun Du, Yuning Du, Lingfeng Zhu, Baohua Lai, Xiaoguang Hu, et~al.
\newblock Pp-ocrv3: More attempts for the improvement of ultra lightweight ocr system.
\newblock \emph{arXiv:2206.03001}, 2022.

\bibitem[Li et~al.(2024{\natexlab{a}})Li, Wang, Xu, Wang, Feng, Kong, and Liu]{li2024multimodal_arxivQA}
Lei Li, Yuqi Wang, Runxin Xu, Peiyi Wang, Xiachong Feng, Lingpeng Kong, and Qi~Liu.
\newblock Multimodal arxiv: A dataset for improving scientific comprehension of large vision-language models.
\newblock \emph{arXiv:2403.00231}, 2024{\natexlab{a}}.

\bibitem[Li et~al.(2024{\natexlab{b}})Li, Lin, Chiang, Weinman, Tual, Chazalon, Perret, Dum{\'e}nieu, and Abadie]{li2024icdar_maptext}
Zekun Li, Yijun Lin, Yao-Yi Chiang, Jerod Weinman, Solenn Tual, Joseph Chazalon, Julien Perret, Bertrand Dum{\'e}nieu, and Nathalie Abadie.
\newblock {ICDAR} 2024 competition on historical map text detection, recognition, and linking.
\newblock In \emph{ICDAR}, 2024{\natexlab{b}}.

\bibitem[Li et~al.(2023)Li, Wang, Stengel-Eskin, Kortylewski, Ma, Van~Durme, and Yuille]{li2023superclevr}
Zhuowan Li, Xingrui Wang, Elias Stengel-Eskin, Adam Kortylewski, Wufei Ma, Benjamin Van~Durme, and Alan~L Yuille.
\newblock Super-clevr: A virtual benchmark to diagnose domain robustness in visual reasoning.
\newblock In \emph{CVPR}, pp.\  14963--14973, 2023.

\bibitem[Lindstr{\"o}m \& Abraham(2022)Lindstr{\"o}m and Abraham]{lindstrom2022clevrmath}
Adam~Dahlgren Lindstr{\"o}m and Savitha~Sam Abraham.
\newblock Clevr-math: A dataset for compositional language, visual and mathematical reasoning.
\newblock \emph{arXiv preprint arXiv:2208.05358}, 2022.

\bibitem[Liu et~al.(2023{\natexlab{a}})Liu, Lin, Li, Wang, Yacoob, and Wang]{liu2023lrv-instruction}
Fuxiao Liu, Kevin Lin, Linjie Li, Jianfeng Wang, Yaser Yacoob, and Lijuan Wang.
\newblock Aligning large multi-modal model with robust instruction tuning.
\newblock \emph{arXiv preprint arXiv:2306.14565}, 2023{\natexlab{a}}.

\bibitem[Liu et~al.(2023{\natexlab{b}})Liu, Wang, Yao, Chen, Song, Cho, Yacoob, and Yu]{liu2023mmcinst}
Fuxiao Liu, Xiaoyang Wang, Wenlin Yao, Jianshu Chen, Kaiqiang Song, Sangwoo Cho, Yaser Yacoob, and Dong Yu.
\newblock {MMC}: Advancing multimodal chart understanding with large-scale instruction tuning.
\newblock \emph{arXiv:2311.10774}, 2023{\natexlab{b}}.

\bibitem[Liu et~al.(2023{\natexlab{c}})Liu, Li, Li, and Lee]{llava1.5}
Haotian Liu, Chunyuan Li, Yuheng Li, and Yong~Jae Lee.
\newblock Improved baselines with visual instruction tuning.
\newblock \emph{CoRR}, abs/2310.03744, 2023{\natexlab{c}}.

\bibitem[Liu et~al.(2023{\natexlab{d}})Liu, Li, Wu, and Lee]{liu2023llava}
Haotian Liu, Chunyuan Li, Qingyang Wu, and Yong~Jae Lee.
\newblock Visual instruction tuning.
\newblock 36, 2023{\natexlab{d}}.

\bibitem[Liu et~al.(2023{\natexlab{e}})Liu, Li, Wu, and Lee]{llava}
Haotian Liu, Chunyuan Li, Qingyang Wu, and Yong~Jae Lee.
\newblock Visual instruction tuning.
\newblock \emph{CoRR}, abs/2304.08485, 2023{\natexlab{e}}.

\bibitem[Liu et~al.(2024{\natexlab{a}})Liu, Cao, Gao, Wang, Chen, Wang, Tian, Lu, Zhu, Lu, et~al.]{liu2024mminstruct}
Yangzhou Liu, Yue Cao, Zhangwei Gao, Weiyun Wang, Zhe Chen, Wenhai Wang, Hao Tian, Lewei Lu, Xizhou Zhu, Tong Lu, et~al.
\newblock Mminstruct: A high-quality multi-modal instruction tuning dataset with extensive diversity.
\newblock \emph{arXiv:2407.15838}, 2024{\natexlab{a}}.

\bibitem[Liu et~al.(2023{\natexlab{f}})Liu, Li, Li, Yu, Huang, Peng, Liu, Chen, Li, Jin, et~al.]{liu2023ocrbench}
Yuliang Liu, Zhang Li, Hongliang Li, Wenwen Yu, Mingxin Huang, Dezhi Peng, Mingyu Liu, Mingrui Chen, Chunyuan Li, Lianwen Jin, et~al.
\newblock On the hidden mystery of ocr in large multimodal models.
\newblock \emph{arXiv preprint arXiv:2305.07895}, 2023{\natexlab{f}}.

\bibitem[Liu et~al.(2023{\natexlab{g}})Liu, Li, Li, Yu, Huang, Peng, Liu, Chen, Li, Jin, et~al.]{llmocr}
Yuliang Liu, Zhang Li, Hongliang Li, Wenwen Yu, Mingxin Huang, Dezhi Peng, Mingyu Liu, Mingrui Chen, Chunyuan Li, Lianwen Jin, et~al.
\newblock On the hidden mystery of ocr in large multimodal models.
\newblock \emph{arXiv preprint arXiv:2305.07895}, 2023{\natexlab{g}}.

\bibitem[Liu et~al.(2024{\natexlab{b}})Liu, Chu, Zang, Wei, Dong, Zhang, Liang, Xiong, Qiao, Lin, et~al.]{liu2024mmdu}
Ziyu Liu, Tao Chu, Yuhang Zang, Xilin Wei, Xiaoyi Dong, Pan Zhang, Zijian Liang, Yuanjun Xiong, Yu~Qiao, Dahua Lin, et~al.
\newblock Mmdu: A multi-turn multi-image dialog understanding benchmark and instruction-tuning dataset for lvlms.
\newblock \emph{arXiv:2406.11833}, 2024{\natexlab{b}}.

\bibitem[Lu et~al.(2021{\natexlab{a}})Lu, Gong, Jiang, Qiu, Huang, Liang, and Zhu]{lu2021geometry3k}
Pan Lu, Ran Gong, Shibiao Jiang, Liang Qiu, Siyuan Huang, Xiaodan Liang, and Song-Chun Zhu.
\newblock Inter-gps: Interpretable geometry problem solving with formal language and symbolic reasoning.
\newblock \emph{arXiv preprint arXiv:2105.04165}, 2021{\natexlab{a}}.

\bibitem[Lu et~al.(2021{\natexlab{b}})Lu, Qiu, Chen, Xia, Zhao, Zhang, Yu, Liang, and Zhu]{lu2021iconqa}
Pan Lu, Liang Qiu, Jiaqi Chen, Tony Xia, Yizhou Zhao, Wei Zhang, Zhou Yu, Xiaodan Liang, and Song-Chun Zhu.
\newblock Iconqa: A new benchmark for abstract diagram understanding and visual language reasoning.
\newblock \emph{arXiv preprint arXiv:2110.13214}, 2021{\natexlab{b}}.

\bibitem[Lu et~al.(2022{\natexlab{a}})Lu, Mishra, Xia, Qiu, Chang, Zhu, Tafjord, Clark, and Kalyan]{lu2022scienceqa}
Pan Lu, Swaroop Mishra, Tanglin Xia, Liang Qiu, Kai-Wei Chang, Song-Chun Zhu, Oyvind Tafjord, Peter Clark, and Ashwin Kalyan.
\newblock Learn to explain: Multimodal reasoning via thought chains for science question answering.
\newblock 35:\penalty0 2507--2521, 2022{\natexlab{a}}.

\bibitem[Lu et~al.(2022{\natexlab{b}})Lu, Qiu, Chang, Wu, Zhu, Rajpurohit, Clark, and Kalyan]{lu2022tablemwp}
Pan Lu, Liang Qiu, Kai-Wei Chang, Ying~Nian Wu, Song-Chun Zhu, Tanmay Rajpurohit, Peter Clark, and Ashwin Kalyan.
\newblock Dynamic prompt learning via policy gradient for semi-structured mathematical reasoning.
\newblock \emph{arXiv preprint arXiv:2209.14610}, 2022{\natexlab{b}}.

\bibitem[Lu et~al.(2024)Lu, Jiang, Chen, Wang, Choi, and Lin]{lu2024wildvision}
Yujie Lu, Dongfu Jiang, Wenhu Chen, William~Yang Wang, Yejin Choi, and Bill~Yuchen Lin.
\newblock Wildvision: Evaluating vision-language models in the wild with human preferences.
\newblock \emph{arXiv:2406.11069}, 2024.

\bibitem[Mahamoud et~al.(2024)Mahamoud, Coustaty, Joseph, d’Andecy, and Ogier]{mahamoud2024chic_vqa_cd}
Ibrahim~Souleiman Mahamoud, Micka{\"e}l Coustaty, Aur{\'e}lie Joseph, Vincent~Poulain d’Andecy, and Jean-Marc Ogier.
\newblock Chic: Corporate document for visual question answering.
\newblock In \emph{ICDAR}, 2024.

\bibitem[Marti \& Bunke(2002)Marti and Bunke]{marti2002iam}
U-V Marti and Horst Bunke.
\newblock The iam-database: an english sentence database for offline handwriting recognition.
\newblock \emph{International Journal on Document Analysis and Recognition}, 5:\penalty0 39--46, 2002.

\bibitem[Masry et~al.(2022{\natexlab{a}})Masry, Do, Tan, Joty, and Hoque]{masry2022chartqa}
Ahmed Masry, Xuan~Long Do, Jia~Qing Tan, Shafiq Joty, and Enamul Hoque.
\newblock Chartqa: A benchmark for question answering about charts with visual and logical reasoning.
\newblock pp.\  2263--2279, 2022{\natexlab{a}}.

\bibitem[Masry et~al.(2022{\natexlab{b}})Masry, Long, Tan, Joty, and Hoque]{chartqa}
Ahmed Masry, Do~Xuan Long, Jia~Qing Tan, Shafiq~R. Joty, and Enamul Hoque.
\newblock Chartqa: {A} benchmark for question answering about charts with visual and logical reasoning.
\newblock In \emph{{ACL} (Findings)}, pp.\  2263--2279. Association for Computational Linguistics, 2022{\natexlab{b}}.

\bibitem[Masry et~al.(2023)Masry, Kavehzadeh, Do, Hoque, and Joty]{masry2023unichart}
Ahmed Masry, Parsa Kavehzadeh, Xuan~Long Do, Enamul Hoque, and Shafiq Joty.
\newblock {UniChart}: A universal vision-language pretrained model for chart comprehension and reasoning.
\newblock \emph{arXiv:2305.14761}, 2023.

\bibitem[Mathew et~al.(2021)Mathew, Karatzas, and Jawahar]{docvqa}
Minesh Mathew, Dimosthenis Karatzas, and C.~V. Jawahar.
\newblock Docvqa: {A} dataset for {VQA} on document images.
\newblock In \emph{{WACV}}, pp.\  2199--2208. {IEEE}, 2021.

\bibitem[Mathew et~al.(2022{\natexlab{a}})Mathew, Bagal, Tito, Karatzas, Valveny, and Jawahar]{infovqa}
Minesh Mathew, Viraj Bagal, Rub{\`{e}}n Tito, Dimosthenis Karatzas, Ernest Valveny, and C.~V. Jawahar.
\newblock Infographicvqa.
\newblock In \emph{{WACV}}, pp.\  2582--2591. {IEEE}, 2022{\natexlab{a}}.

\bibitem[Mathew et~al.(2022{\natexlab{b}})Mathew, Bagal, Tito, Karatzas, Valveny, and Jawahar]{mathew2022infographicvqa}
Minesh Mathew, Viraj Bagal, Rub{\`e}n Tito, Dimosthenis Karatzas, Ernest Valveny, and CV~Jawahar.
\newblock Infographicvqa.
\newblock pp.\  1697--1706, 2022{\natexlab{b}}.

\bibitem[Methani et~al.(2020)Methani, Ganguly, Khapra, and Kumar]{methani2020plotqa}
Nitesh Methani, Pritha Ganguly, Mitesh~M Khapra, and Pratyush Kumar.
\newblock Plotqa: Reasoning over scientific plots.
\newblock pp.\  1527--1536, 2020.

\bibitem[Mishra et~al.(2012)Mishra, Alahari, and Jawahar]{mishra2012scene_iiit5k}
Anand Mishra, Karteek Alahari, and CV~Jawahar.
\newblock Scene text recognition using higher order language priors.
\newblock In \emph{BMVC}, 2012.

\bibitem[Mishra et~al.(2019)Mishra, Shekhar, Singh, and Chakraborty]{mishra2019ocrvqa}
Anand Mishra, Shashank Shekhar, Ajeet~Kumar Singh, and Anirban Chakraborty.
\newblock Ocr-vqa: Visual question answering by reading text in images.
\newblock pp.\  947--952, 2019.

\bibitem[mychen76(2024)]{mychen76_invoices_receipts_ocr_v1}
mychen76.
\newblock Invoices and receipts ocr v1 dataset.
\newblock \url{https://huggingface.co/datasets/mychen76/invoices-and-receipts_ocr_v1}, 2024.

\bibitem[OleehyO(2024)]{oleehyo_latex_formulas}
OleehyO.
\newblock Latex formulas dataset.
\newblock \url{https://huggingface.co/datasets/OleehyO/latex-formulas}, 2024.

\bibitem[OpenAI(2024)]{openai2024gpt4o}
OpenAI.
\newblock {Hello GPT-4o}.
\newblock \url{https://openai.com/index/gpt-4v-system-card}, 2024.
\newblock Accessed: 2024-12-29.

\bibitem[parasam(2024)]{captcha}
parasam.
\newblock Captcha dataset.
\newblock \url{https://www.kaggle.com/datasets/parsasam/captcha-dataset}, 2024.

\bibitem[Radford et~al.(2021)Radford, Kim, Hallacy, Ramesh, Goh, Agarwal, Sastry, Askell, Mishkin, Clark, Krueger, and Sutskever]{clip}
Alec Radford, Jong~Wook Kim, Chris Hallacy, Aditya Ramesh, Gabriel Goh, Sandhini Agarwal, Girish Sastry, Amanda Askell, Pamela Mishkin, Jack Clark, Gretchen Krueger, and Ilya Sutskever.
\newblock Learning transferable visual models from natural language supervision.
\newblock In \emph{{ICML}}, volume 139 of \emph{Proceedings of Machine Learning Research}, pp.\  8748--8763. {PMLR}, 2021.

\bibitem[Schwenk et~al.(2022)Schwenk, Khandelwal, Clark, Marino, and Mottaghi]{schwenk2022aokvqa}
Dustin Schwenk, Apoorv Khandelwal, Christopher Clark, Kenneth Marino, and Roozbeh Mottaghi.
\newblock A-okvqa: A benchmark for visual question answering using world knowledge.
\newblock In \emph{ECCV}, pp.\  146--162, 2022.

\bibitem[Seo et~al.(2015)Seo, Hajishirzi, Farhadi, Etzioni, and Malcolm]{seo2015geos}
Minjoon Seo, Hannaneh Hajishirzi, Ali Farhadi, Oren Etzioni, and Clint Malcolm.
\newblock Solving geometry problems: Combining text and diagram interpretation.
\newblock In \emph{EMNLP}, 2015.

\bibitem[Singh et~al.(2019{\natexlab{a}})Singh, Natarajan, Shah, Jiang, Chen, Batra, Parikh, and Rohrbach]{singh2019textvqa}
Amanpreet Singh, Vivek Natarajan, Meet Shah, Yu~Jiang, Xinlei Chen, Dhruv Batra, Devi Parikh, and Marcus Rohrbach.
\newblock Towards vqa models that can read.
\newblock In \emph{CVPR}, pp.\  8317--8326, 2019{\natexlab{a}}.

\bibitem[Singh et~al.(2019{\natexlab{b}})Singh, Natarajan, Shah, Jiang, Chen, Batra, Parikh, and Rohrbach]{textvqa}
Amanpreet Singh, Vivek Natarajan, Meet Shah, Yu~Jiang, Xinlei Chen, Dhruv Batra, Devi Parikh, and Marcus Rohrbach.
\newblock Towards {VQA} models that can read.
\newblock In \emph{{CVPR}}, pp.\  8317--8326. Computer Vision Foundation / {IEEE}, 2019{\natexlab{b}}.

\bibitem[Sujet~AI(2024)]{Sujet-Finance-QA-Vision-100k}
Hamed~Rahimi Sujet~AI, Allaa~Boutaleb.
\newblock Sujet-finance-qa-vision-100k: A large-scale dataset for financial document vqa.
\newblock https://huggingface.co/datasets/sujet-ai/Sujet-Finance-QA-Vision-100k, 2024.

\bibitem[Sun et~al.(2019)Sun, Ni, Chng, Liu, Luo, Ng, Han, Ding, Liu, Karatzas, et~al.]{sun2019lsvt}
Yipeng Sun, Zihan Ni, Chee-Kheng Chng, Yuliang Liu, Canjie Luo, Chun~Chet Ng, Junyu Han, Errui Ding, Jingtuo Liu, Dimosthenis Karatzas, et~al.
\newblock Icdar 2019 competition on large-scale street view text with partial labeling-rrc-lsvt.
\newblock pp.\  1557--1562, 2019.

\bibitem[TAL(2023)]{tal}
TAL.
\newblock Tal open dataset.
\newblock \url{https://ai.100tal.com/dataset}, 2023.

\bibitem[Tang et~al.(2024)Tang, Liu, Ye, Lu, Wei, Lin, Li, Mahmood, Feng, Zhao, et~al.]{tang2024mtvqa}
Jingqun Tang, Qi~Liu, Yongjie Ye, Jinghui Lu, Shu Wei, Chunhui Lin, Wanqing Li, Mohamad Fitri Faiz~Bin Mahmood, Hao Feng, Zhen Zhao, et~al.
\newblock Mtvqa: Benchmarking multilingual text-centric visual question answering.
\newblock \emph{arXiv preprint arXiv:2405.11985}, 2024.

\bibitem[Team et~al.(2023)Team, Anil, Borgeaud, Wu, Alayrac, Yu, Soricut, Schalkwyk, Dai, Hauth, et~al.]{team2023gemini}
Gemini Team, Rohan Anil, Sebastian Borgeaud, Yonghui Wu, Jean-Baptiste Alayrac, Jiahui Yu, Radu Soricut, Johan Schalkwyk, Andrew~M Dai, Anja Hauth, et~al.
\newblock Gemini: a family of highly capable multimodal models.
\newblock \emph{arXiv preprint arXiv:2312.11805}, 2023.

\bibitem[Team(2024)]{chen2024internvl2}
OpenGVLab Team.
\newblock {InternVL2}: Better than the best—expanding performance boundaries of open-source multimodal models with the progressive scaling strategy.
\newblock \url{https://internvl.github.io/blog/2024-07-02-InternVL-2.0/}, 2024.

\bibitem[TIGER-Lab(2024)]{tiger_lab_visualwebinstruct}
TIGER-Lab.
\newblock Visualwebinstruct dataset.
\newblock \url{https://huggingface.co/datasets/TIGER-Lab/VisualWebInstruct}, 2024.

\bibitem[Touvron et~al.(2023)Touvron, Lavril, Izacard, Martinet, Lachaux, Lacroix, Rozi{\`e}re, Goyal, Hambro, Azhar, et~al.]{llama}
Hugo Touvron, Thibaut Lavril, Gautier Izacard, Xavier Martinet, Marie-Anne Lachaux, Timoth{\'e}e Lacroix, Baptiste Rozi{\`e}re, Naman Goyal, Eric Hambro, Faisal Azhar, et~al.
\newblock Llama: Open and efficient foundation language models.
\newblock \emph{arXiv preprint arXiv:2302.13971}, 2023.

\bibitem[Veit et~al.(2016)Veit, Matera, Neumann, Matas, and Belongie]{veit2016cocotext}
Andreas Veit, Tomas Matera, Lukas Neumann, Jiri Matas, and Serge Belongie.
\newblock {COCO-Text}: Dataset and benchmark for text detection and recognition in natural images.
\newblock \emph{arXiv:1601.07140}, 2016.

\bibitem[Vicuna(2023)]{vicuna}
Vicuna.
\newblock Vicuna: An open chatbot impressing gpt-4.
\newblock \url{https://github.com/lm-sys/FastChat}, 2023.

\bibitem[Wang et~al.(2023)Wang, Meng, Weng, He, Wu, and Jiang]{wang2023lvisinstruct4v}
Junke Wang, Lingchen Meng, Zejia Weng, Bo~He, Zuxuan Wu, and Yu-Gang Jiang.
\newblock To see is to believe: Prompting gpt-4v for better visual instruction tuning.
\newblock \emph{arXiv preprint arXiv:2311.07574}, 2023.

\bibitem[Wang et~al.(2024)Wang, Bai, Tan, Wang, Fan, Bai, Chen, Liu, Wang, Ge, et~al.]{wang2024qwen2vl}
Peng Wang, Shuai Bai, Sinan Tan, Shijie Wang, Zhihao Fan, Jinze Bai, Keqin Chen, Xuejing Liu, Jialin Wang, Wenbin Ge, et~al.
\newblock Qwen2-vl: Enhancing vision-language model's perception of the world at any resolution.
\newblock \emph{arXiv preprint arXiv:2409.12191}, 2024.

\bibitem[Wang et~al.(2020)Wang, Liu, Shen, Ng, Luo, Jin, Chan, Hengel, and Wang]{wang2020estvqa}
Xinyu Wang, Yuliang Liu, Chunhua Shen, Chun~Chet Ng, Canjie Luo, Lianwen Jin, Chee~Seng Chan, Anton van~den Hengel, and Liangwei Wang.
\newblock On the general value of evidence, and bilingual scene-text visual question answering.
\newblock In \emph{CVPR}, 2020.

\bibitem[Wu et~al.(2024)Wu, Zhu, Bai, Liang, Li, Wu, Liu, Liu, Qu, Cheng, et~al.]{wu2024mmra}
Siwei Wu, Kang Zhu, Yu~Bai, Yiming Liang, Yizhi Li, Haoning Wu, Jiaheng Liu, Ruibo Liu, Xingwei Qu, Xuxin Cheng, et~al.
\newblock Mmra: A benchmark for multi-granularity multi-image relational association.
\newblock \emph{arXiv:2407.17379}, 2024.

\bibitem[Xie et~al.(2022)Xie, Fu, Zhang, Wang, and Bai]{xie2022toward_wordart}
Xudong Xie, Ling Fu, Zhifei Zhang, Zhaowen Wang, and Xiang Bai.
\newblock Toward understanding wordart: Corner-guided transformer for scene text recognition.
\newblock In \emph{ECCV}, 2022.

\bibitem[Xiong et~al.(2024)Xiong, Wang, Guo, Ye, Fan, Gu, Huang, and Li]{xiong2024llava_critic}
Tianyi Xiong, Xiyao Wang, Dong Guo, Qinghao Ye, Haoqi Fan, Quanquan Gu, Heng Huang, and Chunyuan Li.
\newblock Llava-critic: Learning to evaluate multimodal models.
\newblock \emph{arXiv:2410.02712}, 2024.

\bibitem[Xu et~al.(2023)Xu, Sun, Zheng, Geng, Zhao, Feng, Tao, and Jiang]{xu2023wizardlm}
Can Xu, Qingfeng Sun, Kai Zheng, Xiubo Geng, Pu~Zhao, Jiazhan Feng, Chongyang Tao, and Daxin Jiang.
\newblock Wizardlm: Empowering large language models to follow complex instructions.
\newblock \emph{arXiv:2304.12244}, 2023.

\bibitem[Yao et~al.(2024)Yao, Yu, Zhang, Wang, Cui, Zhu, Cai, Li, Zhao, He, et~al.]{yao2024minicpm}
Yuan Yao, Tianyu Yu, Ao~Zhang, Chongyi Wang, Junbo Cui, Hongji Zhu, Tianchi Cai, Haoyu Li, Weilin Zhao, Zhihui He, et~al.
\newblock Minicpm-v: A gpt-4v level mllm on your phone.
\newblock \emph{arXiv preprint arXiv:2408.01800}, 2024.

\bibitem[Ye et~al.(2023)Ye, Hu, Xu, Ye, Yan, Xu, Li, Tian, Qian, Zhang, et~al.]{ye2023ureader}
Jiabo Ye, Anwen Hu, Haiyang Xu, Qinghao Ye, Ming Yan, Guohai Xu, Chenliang Li, Junfeng Tian, Qi~Qian, Ji~Zhang, et~al.
\newblock Ureader: Universal ocr-free visually-situated language understanding with multimodal large language model.
\newblock \emph{arXiv preprint arXiv:2310.05126}, 2023.

\bibitem[Yu et~al.(2023{\natexlab{a}})Yu, Jiang, Shi, Yu, Liu, Zhang, Kwok, Li, Weller, and Liu]{yu2023mathqa}
Longhui Yu, Weisen Jiang, Han Shi, Jincheng Yu, Zhengying Liu, Yu~Zhang, James~T Kwok, Zhenguo Li, Adrian Weller, and Weiyang Liu.
\newblock {MetaMath}: Bootstrap your own mathematical questions for large language models.
\newblock \emph{arXiv:2309.12284}, 2023{\natexlab{a}}.

\bibitem[Yu et~al.(2023{\natexlab{b}})Yu, Zhang, Cao, Hua, Li, Chen, Liu, Chen, Kuang, Cheng, et~al.]{yu2023icdar_svrd}
Wenwen Yu, Chengquan Zhang, Haoyu Cao, Wei Hua, Bohan Li, Huang Chen, Mingyu Liu, Mingrui Chen, Jianfeng Kuang, Mengjun Cheng, et~al.
\newblock Icdar 2023 competition on structured text extraction from visually-rich document images.
\newblock In \emph{ICDAR}, 2023{\natexlab{b}}.

\bibitem[Yuan et~al.(2024)Yuan, Cui, Wang, Ding, Wang, Deng, Shan, Chen, Xie, Lin, et~al.]{yuan2024advancing_ultrainteract}
Lifan Yuan, Ganqu Cui, Hanbin Wang, Ning Ding, Xingyao Wang, Jia Deng, Boji Shan, Huimin Chen, Ruobing Xie, Yankai Lin, et~al.
\newblock Advancing llm reasoning generalists with preference trees.
\newblock \emph{arXiv:2404.02078}, 2024.

\bibitem[Yuan et~al.(2019)Yuan, Zhu, Xu, Li, Mu, and Hu]{yuan2019ctw}
Tai-Ling Yuan, Zhe Zhu, Kun Xu, Cheng-Jun Li, Tai-Jiang Mu, and Shi-Min Hu.
\newblock A large chinese text dataset in the wild.
\newblock \emph{Journal of Computer Science and Technology}, 34:\penalty0 509--521, 2019.

\bibitem[Zhang et~al.(2024)Zhang, Wei, Jiang, Zhang, Guo, Tong, Liu, Zhou, Wei, Zhang, et~al.]{zhang2024mavis}
Renrui Zhang, Xinyu Wei, Dongzhi Jiang, Yichi Zhang, Ziyu Guo, Chengzhuo Tong, Jiaming Liu, Aojun Zhou, Bin Wei, Shanghang Zhang, et~al.
\newblock Mavis: Mathematical visual instruction tuning.
\newblock \emph{arXiv preprint arXiv:2407.08739}, 2024.

\bibitem[Zhang et~al.(2019)Zhang, Zhou, Jiang, Song, Li, Zhou, Wang, Wang, Liao, Yang, et~al.]{zhang2019rects}
Rui Zhang, Yongsheng Zhou, Qianyi Jiang, Qi~Song, Nan Li, Kai Zhou, Lei Wang, Dong Wang, Minghui Liao, Mingkun Yang, et~al.
\newblock Icdar 2019 robust reading challenge on reading chinese text on signboard.
\newblock pp.\  1577--1581, 2019.

\bibitem[Zhang et~al.(2023)Zhang, Zhang, Gu, Zhou, Lipka, Yang, and Sun]{zhang2023llavar}
Yanzhe Zhang, Ruiyi Zhang, Jiuxiang Gu, Yufan Zhou, Nedim Lipka, Diyi Yang, and Tong Sun.
\newblock Llavar: Enhanced visual instruction tuning for text-rich image understanding.
\newblock \emph{arXiv preprint arXiv:2306.17107}, 2023.

\bibitem[Zhao et~al.(2023)Zhao, Zhao, Nan, Qi, Zhang, Tang, Mi, and Radev]{zhao2023robut}
Yilun Zhao, Chen Zhao, Linyong Nan, Zhenting Qi, Wenlin Zhang, Xiangru Tang, Boyu Mi, and Dragomir Radev.
\newblock Robut: A systematic study of table qa robustness against human-annotated adversarial perturbations.
\newblock \emph{arXiv:2306.14321}, 2023.

\bibitem[Zhu et~al.(2023)Zhu, Chen, Shen, Li, and Elhoseiny]{minigpt4}
Deyao Zhu, Jun Chen, Xiaoqian Shen, Xiang Li, and Mohamed Elhoseiny.
\newblock Minigpt-4: Enhancing vision-language understanding with advanced large language models, 2023.

\bibitem[Zhu et~al.(2022)Zhu, Lei, Feng, Wang, Zhang, and Chua]{zhu2022tatdqa}
Fengbin Zhu, Wenqiang Lei, Fuli Feng, Chao Wang, Haozhou Zhang, and Tat-Seng Chua.
\newblock Towards complex document understanding by discrete reasoning.
\newblock In \emph{ACMMM}, 2022.

\end{thebibliography}
\bibliographystyle{iclr2025_conference}

\clearpage

\appendix
\section{Appendix}

\begin{figure}[h]
    \centering
    \includegraphics[width=\textwidth]{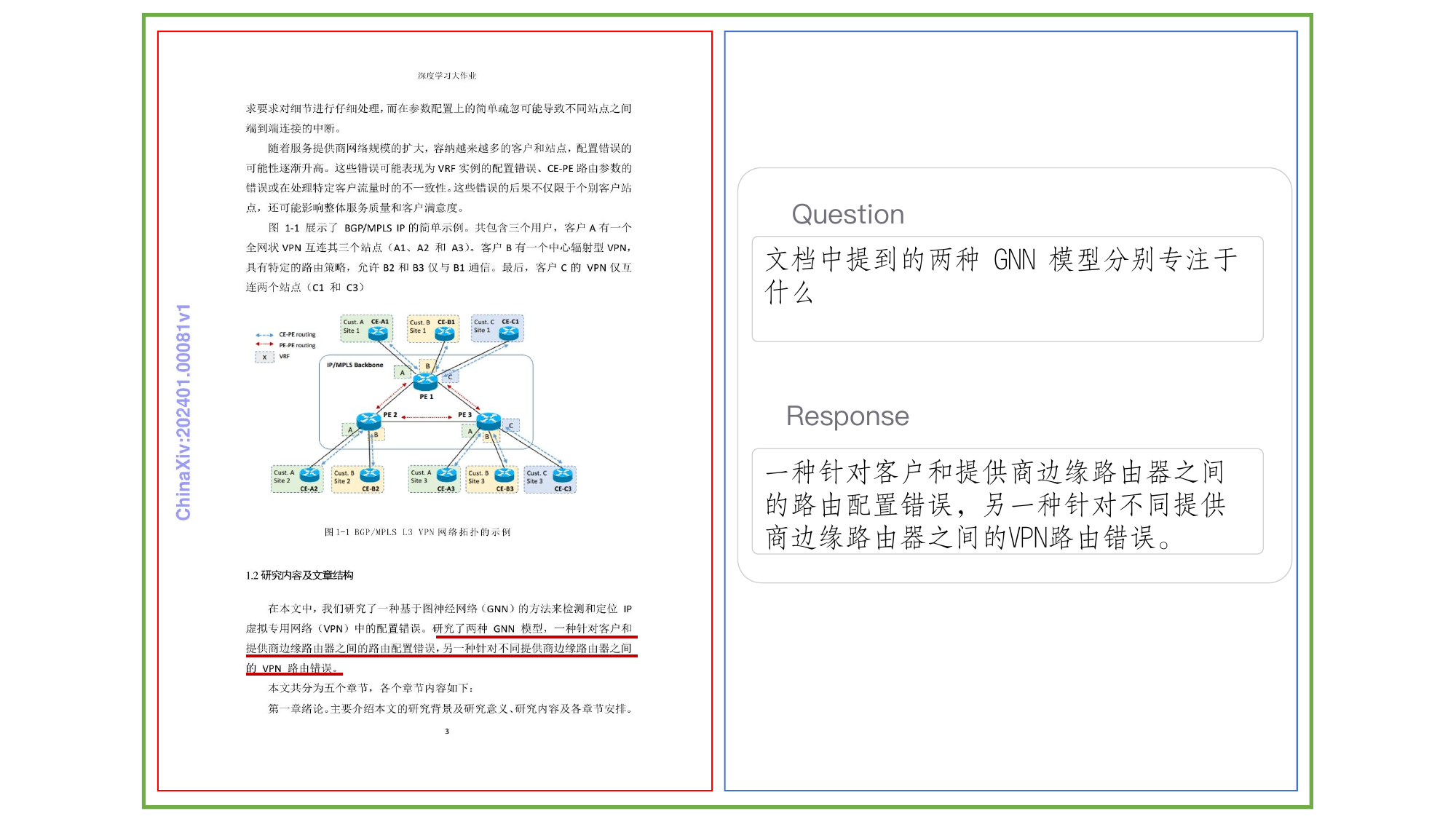}
    \caption{Case in science and engineering papers for document content understanding.}
    \label{fig:case1}
\end{figure}

\begin{figure}[h]
    \centering
    \includegraphics[width=\textwidth]{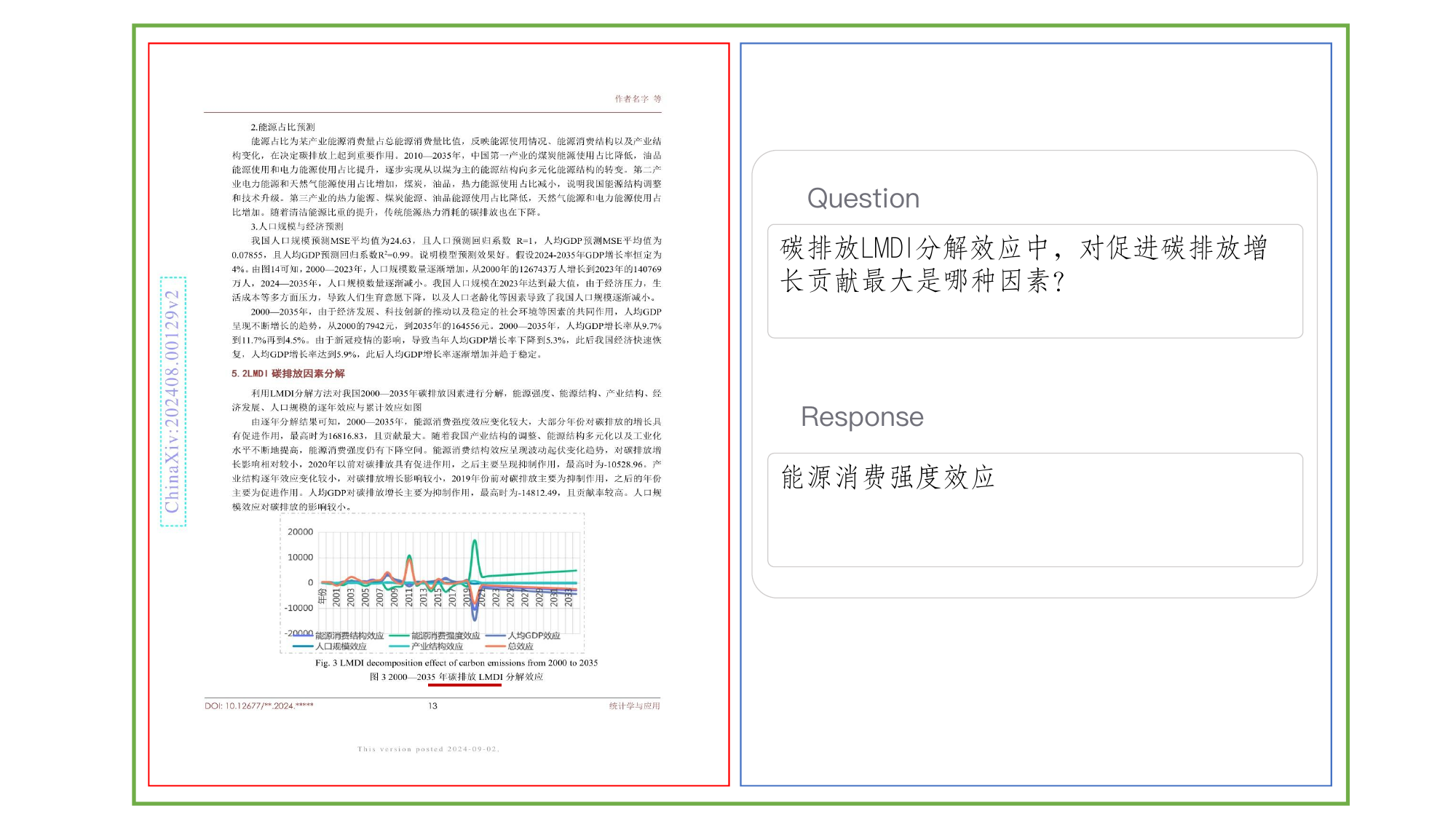}
    \caption{Case in science and engineering papers for table understanding.}
    \label{fig:case2}
\end{figure}

\begin{figure}[h]
    \centering
    \includegraphics[width=\textwidth]{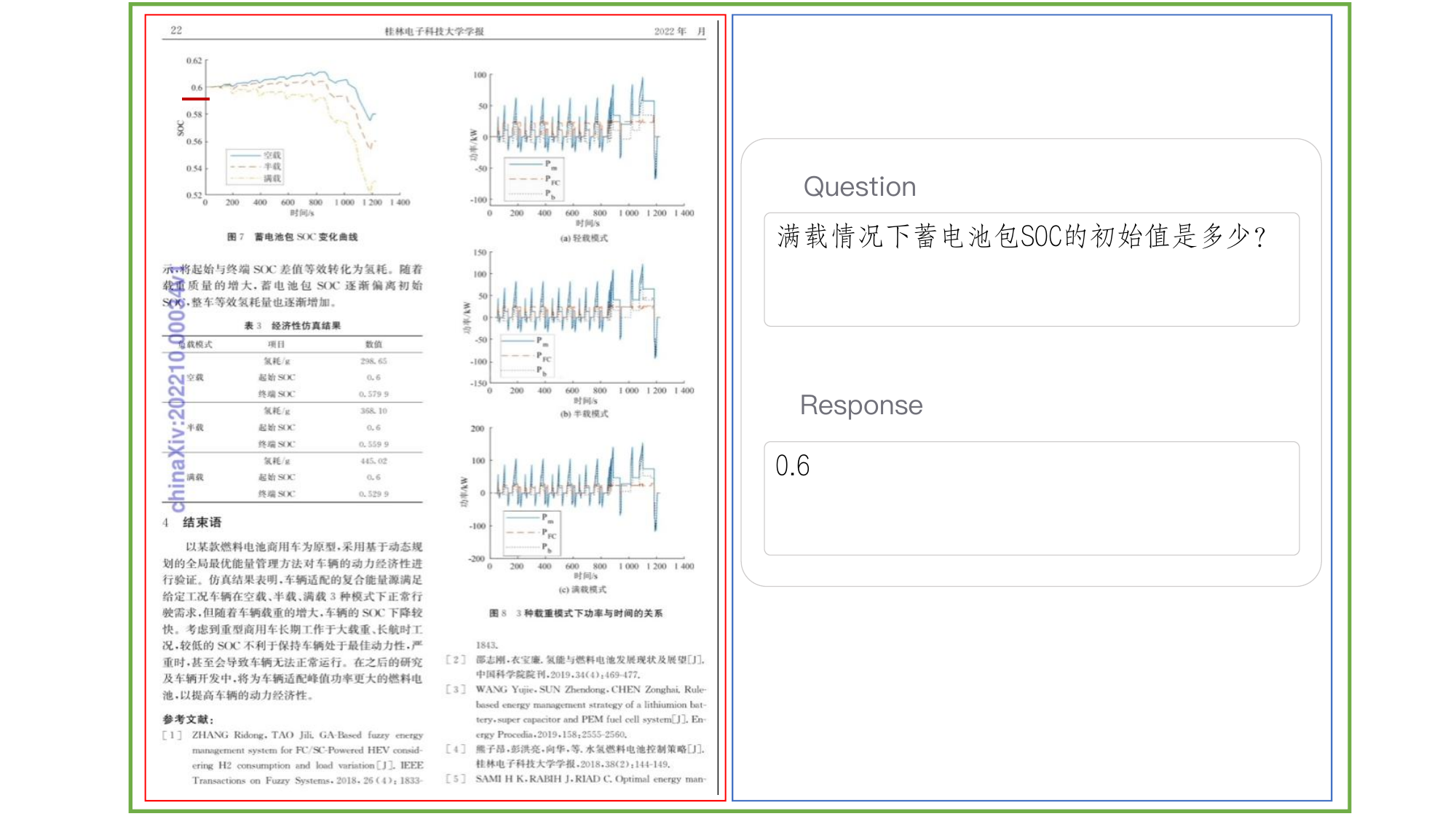}
    \caption{Case in science and engineering papers for chart understanding.}
    \label{fig:case3}
\end{figure}

\begin{figure}[h]
    \centering
    \includegraphics[width=\textwidth]{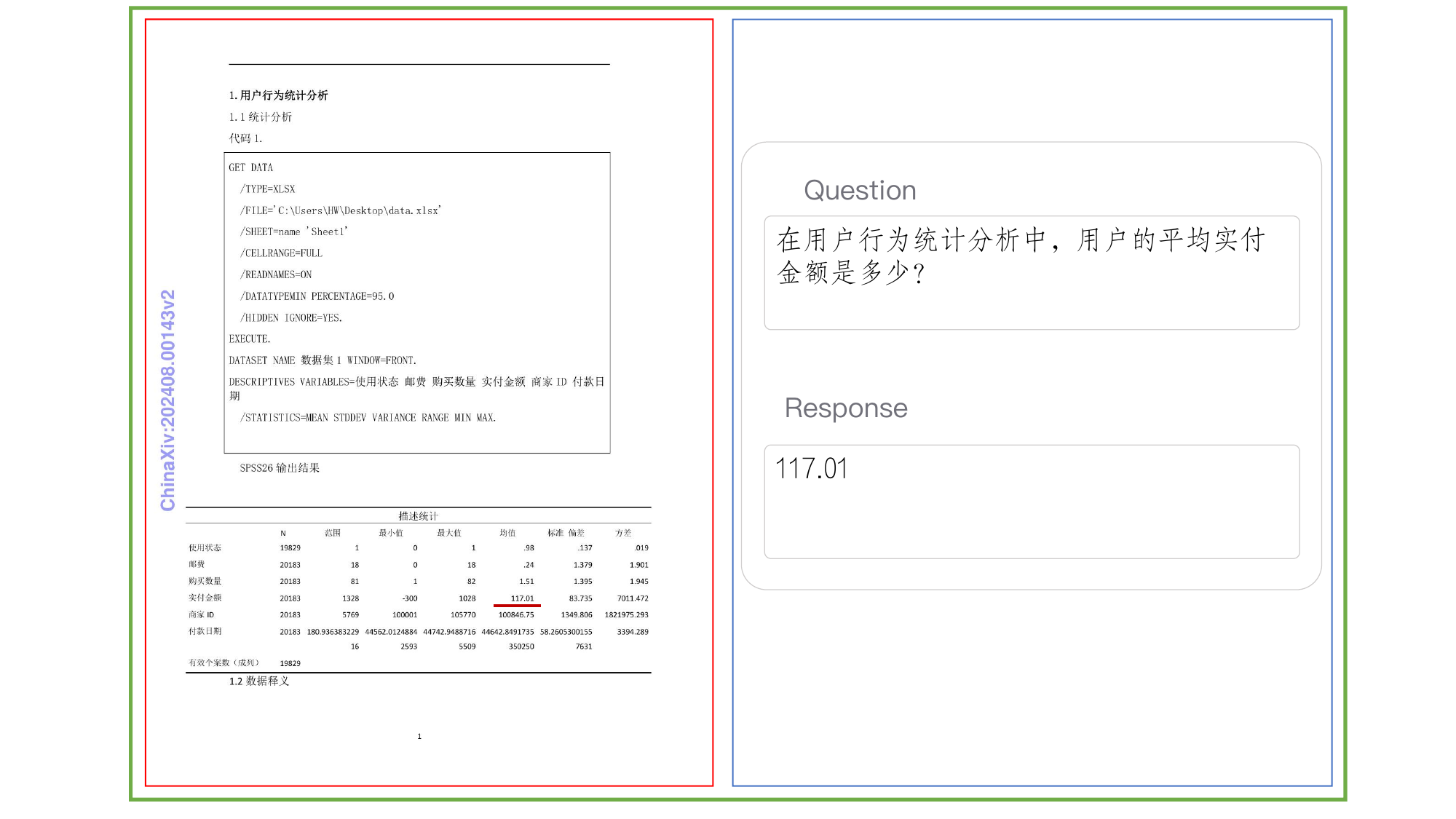}
    \caption{Case in financial reports for table understanding..}
    \label{fig:case4}
\end{figure}

\end{document}